\documentclass[sigconf]{acmart}
\newcommand{\ignore}[1]{}
\usepackage{xspace}

\newcommand{\bm}{\mathbf{m}}

\usepackage{amsfonts,amssymb}
\usepackage{booktabs} 
\usepackage{caption}
\usepackage{subcaption}
\usepackage{bm}
\usepackage{multirow}
\usepackage{footmisc}
\usepackage{amssymb}
\usepackage{array}
\usepackage{graphicx}
\usepackage{clrscode}
\usepackage{balance}
\usepackage{float}
\usepackage{color}
\usepackage{xcolor}
\usepackage{amsopn}
\usepackage{mathrsfs}
\usepackage{mathtools}
\usepackage{amsmath}
\usepackage{arydshln}
\usepackage{hyperref}
\usepackage{blkarray}
\usepackage{enumerate}
\usepackage{courier}
\usepackage{mathrsfs}
\usepackage{rotating}
\usepackage{ragged2e}
\usepackage{multicol}
\usepackage{latexsym}
\usepackage{mflogo}
\usepackage{amsthm}
\usepackage{soul}
\usepackage{hhline}
\usepackage[ruled,linesnumbered]{algorithm2e}
\usepackage{makecell}
\usepackage{bbm}

\usepackage{array}
\usepackage{tabularx}
\usepackage{CJKutf8}
\usepackage[toc, page]{appendix}
\newcommand{\paratitle}[1]{\vspace{1.5ex}\noindent\textbf{#1}}
\newcommand{\ie}{\emph{i.e.,}\xspace}
\newcommand{\aka}{\emph{a.k.a.,}\xspace}
\newcommand{\eg}{\emph{e.g.,}\xspace}
\newcommand{\etal}{\emph{et al.}\xspace}

\AtBeginDocument{%
  \providecommand\BibTeX{{%
    \normalfont B\kern-0.5em{\scshape i\kern-0.25em b}\kern-0.8em\TeX}}}

\copyrightyear{2023}
\acmYear{2023}
\setcopyright{acmlicensed}\acmConference[KDD '23]{Proceedings of the 29th
ACM SIGKDD Conference on Knowledge Discovery and Data Mining}{August
6--10, 2023}{Long Beach, CA, USA}
\acmBooktitle{Proceedings of the 29th ACM SIGKDD Conference on Knowledge
Discovery and Data Mining (KDD '23), August 6--10, 2023, Long Beach, CA,
USA}
\acmPrice{15.00}
\acmDOI{10.1145/3580305.3599850}
\acmISBN{979-8-4007-0103-0/23/08}

\begin{document}

\title{JiuZhang 2.0: A Unified Chinese Pre-trained Language Model for Multi-task Mathematical Problem Solving}

\author{Wayne Xin Zhao$^{\dagger}$, Kun Zhou$^{*}$,  Beichen Zhang$^{*}$}
\thanks{$^{*}$Equal contribution.}
\thanks{$^{\dagger}$Corresponding author.}
\affiliation{%
  \institution{Gaoling School of Artificial Intelligence, School of Information, Renmin University of China}
  \city{Beijing}
  \country{China}
}\email{batmanfly@gmail.com}

\author{Zheng Gong$^{*}$, Zhipeng Chen$^{*}$,  Yuanhang Zhou$^{*}$}
\affiliation{%
  \institution{Gaoling School of Artificial Intelligence, Renmin University of China}
  \city{Beijing}
  \country{China}
}\email{gongzheng0109@ruc.edu.cn}

\author{Ji-Rong Wen}
\affiliation{%
  \institution{Gaoling School of Artificial Intelligence, School of Information, Renmin University of China}
  \city{Beijing}
  \country{China}
}\email{jrwen@ruc.edu.cn}

\author{Jing Sha, Shijin Wang, Cong Liu, Guoping Hu}
\affiliation{%
  \institution{iFLYTEK Research}
\institution{State Key Laboratory of Cognitive Intelligence}
  \city{Hefei}\country{China}\\
  \institution{iFLYTEK AI Research (Central China)}
  \city{Wuhan}
  \country{China}}
\email{jingsha@iflytek.com}
\renewcommand{\shortauthors}{Wayne Xin Zhao et al.}

\begin{abstract} 
Although pre-trained language models~(PLMs) have recently advanced the research progress in mathematical reasoning, they are not specially designed as a capable multi-task solver, suffering from high cost for multi-task deployment (\eg a model copy for a task) and inferior  performance on complex mathematical problems in practical applications. 
To address these issues, in this paper, we propose  \textbf{JiuZhang~2.0}, a unified Chinese PLM specially for multi-task mathematical problem solving. 
Our idea is to maintain a moderate-sized model and employ the \emph{cross-task knowledge sharing} to improve the model capacity in a multi-task setting.  
Specially, we construct a Mixture-of-Experts~(MoE) architecture for modeling mathematical text, so as to capture the common mathematical knowledge across tasks. 
For optimizing   the MoE architecture,  we design \emph{multi-task continual pre-training} 
and \emph{multi-task fine-tuning} strategies for multi-task adaptation.  
These training strategies can effectively decompose the knowledge from the task data and establish the cross-task sharing via expert networks. 
In order to further improve the general capacity of solving different complex tasks, we leverage large language models~(LLMs) as complementary models to iteratively refine the generated solution by our PLM, via  in-context learning.
Extensive experiments have demonstrated the effectiveness of our model.
\end{abstract}




\maketitle

\section{Introduction}
\label{sec:intro}

Recently,  the mathematical reasoning capacity of machines has been largely empowered by the progress of pre-trained language models~(PLMs)~\cite{peng2021mathbert,zhao2022jiuzhang,Lewkowycz2022SolvingQR,Mishra2022LilaAU}.
By pre-training on large-scale mathematical corpus with specially designed tasks, PLMs can understand the mathematical formulas and logic to a certain extent~\cite{zhao2022jiuzhang}, achieving better performance on a variety of math-related tasks. 

\ignore{
Mathematical reasoning~\cite{} is the fundamental ability of mathematical intelligence for machines to understand and handle numerical data and mathematical texts, which requires a wide spectrum of mathematical knowledge, logic and skills.
A surge of works have been proposed in the field of natural language process~(NLP), and incorporate statistical features~\cite{} semantic parser~\cite{}, neural networks~\cite{} and \etal to analyze and understand these mathematical texts.
Recently, pre-trained language models~(PLMs)~\cite{} are also adopted for mathematical reasoning.
By pre-training on large-scale math-related corpus and using specially designed tasks, these PLMs can understand the complex mathematical formulas and logic to a certain extent~\cite{}, achieving better performance on a variety of mathematical problem understanding tasks~\cite{}. 
}

Despite the progress, existing PLM based approaches still have two major limitations  in real-world math-related applications. (1) \emph{Limited task performance}: due to the limit of model capacity and pre-training data, 
PLMs are less capable of understanding  complex mathematical problems, thus suffering  from performance degradation on difficult tasks.  (2) \emph{Large maintenance  cost}:  an online application often supports multiple math-related tasks (\eg  similar problem retrieval and knowledge point classification),  while PLMs   need to be fine-tuned task by task when dealing with different downstream tasks, taking a significant cost of maintaining multi-task solvers (\eg a model copy for a task).  

By exploring the scaling laws,  large language models~(LLMs)\footnote{In this paper,  PLMs and LLMs  refer to mathematical language models with \emph{moderate sizes} (\eg BERT~\cite{devlin2018bert}) and \emph{huge sizes} (\eg GPT-3~\cite{brown2020language}), respectively. }~\cite{brown2020language,chen2021evaluating} can overcome the above issues to some extent with stronger mathematical reasoning ability. While,  they are very costly to be tuned for task or domain adaptation. Although in-context learning~\cite{brown2020language} can be applied to solve different tasks in an efficient way (with no need for fine-tuning), it is still difficult to adapt them to specific tasks that require rich domain knowledge,  \eg English-focused LLMs such as GPT-3~\cite{brown2020language} and CodeX~\cite{chen2021evaluating} cannot perform very well on Chinese mathematical problems (as shown in Table~\ref{tab-hard-results}).  


\ignore{
mathematical problem solving, especially in a multi-task setting.  
Unlike general natural language tasks, different mathematics problems often require specific 
mathematical knowledge for task solving (\eg different operation rules for linear algebra and calculus). While, PLMs implicitly encode the mathematical knowledge via a whole model, and it is difficult to decouple and use the correct knowledge when simultaneously solving multiple  mathematics tasks. Besides, although these PLMs are expected to be multi-task solvers, they are typically fine-fined according to each individual task, lacking sufficient consideration of inter-task relations (\eg the knowledge sharing across two related mathematics tasks). By leveraging a huge network architecture pre-trained on very large corpus, LLMs can overcome the above issues to some extent with stronger mathematical reasoning ability. Since LLMs are very costly to be trained,  it is only feasible to make adaptation for new tasks via efficient strategies (\ie in-context learning~\cite{brown2020language}).
However, in-context learning cannot provide sufficient guidance to LLMs when the setting of  downstream tasks is different from how it is trained. For example, 
we empirically find that  English-focused LLMs such as GPT-3~\cite{brown2020language} and CodeX~\cite{chen2021evaluating} cannot perform very well on Chinese mathematics tasks. 
}

Considering the above issues, we aim to develop a more effective  Chinese PLM that can
well adapt to  multiple complex mathematical tasks, so as to better support math-related applications. 
To motivate our solution, we  observe that mathematical tasks usually rely on common or related background knowledge, \eg a multi-choice problem and a blank-filling problem  might target the same knowledge point though with different problem settings.
Thus, it is intuitive to transfer and share  mathematical  knowledge across tasks by learning a unified model, so that the performance of each individual task can be potentially improved.  In a multi-task manner, it also naturally reduces the cost of task-specific fine-tuning, since  a joint model is trained with the data of all tasks.  
While, to become multi-task learner, it requires a higher generalization ability for solving different tasks~\cite{radford2019language,brown2020language}. For this purpose, we further leverage existing LLMs that implicitly encode large amounts of knowledge to enhance the  capacity of complex problem solving  for PLMs.  


\ignore{
  Indeed, a multi-task learning approach is potentially more superior to address the aforementioned issues of PLMs, \ie limited model capacity and large fine-tuning cost. For model capacity, it transfers useful knowledge across related tasks,  so that the capacity of  solving specific tasks can be improved by leveraging shared task knowledge. In the mathematical applications,  different mathematical tasks usually rely on similar or related background knowledge, \eg multi-choice problems and blank-filling problems might target the same knowledge point though with different problem settings. Thus, we can naturally consider building a unified model that fits multiple tasks (with a small number of task adaptation parameters). 
 For fine-tuning cost, since it aims to learn a unified model to solve multiple tasks at the same time, it naturally reduces the maintenance cost of task-specific fine-tuning. 
}


 
\ignore{
 solve complex mathematical problems in a multi-task setting.  Our major improvements are twofold. Firstly, we utilize the Mixture-of-Experts~(MoE)~\cite{Jacobs1991AdaptiveMO} extension to adapt the PLM to the multi-task setting, and further propose multi-task training methods that are suited to the MoE-based architecture. Such a design can better decouple and share mathematical knowledge across tasks. Second, we utilize a LLM as the complementary model to improve the solution from our PLM.  With this strategy, we can utilize moderate-sized PLM for domain adaption (\ie Chinese mathematical corpus) \emph{with affordable costs},  while employing high-capacity LLM for improving the result of our adapted PLM \emph{with accurate refinement}, combining the merits of both  kinds of language models. 
 }

\ignore{
Despite the success of PLMs, they still perform not well on complex mathematical problem understanding tasks, \eg high-school multi-choice questions and proof problems~\cite{}.
The major reason is that existing PLMs can only rely on large-scale crawled math-related texts, where necessary annotations about mathematical knowledge points and reasoning logic are commonly incomplete or even unavailable.
As a result, although specially designed pre-training tasks have been adopted, it is still hard for these PLMs to systematically learn complex mathematical logic and theorems as human.
When fine-tuning on more complex math-related tasks, they might utilize improper knowledge points by mistake when solving mathematical problems, and generate incorrect computation process (\eg $10\times99=900$) that contradicts the mathematical reasoning logic.
}

\ignore{
To address the above issues, the key is to improve the understanding of mathematical knowledge points and reasoning logic for PLMs.
However, as discussed before, such important information can not be sufficiently acquired from the pre-training corpus.
There are also several works that greatly increase the amount of training data and the number of PLM parameters into billion scale.
Although such a way can enhance the reasoning capacity of PLMs in solving mathematical problems to some extent, it is also hard to collect the required large amount of math-related data for pre-training, especially for collecting a large-scale Chinese math-related corpus.
Besides, the training cost of the large PLM is also unaffordable for real-world applications. 
Therefore, in this work, we focus on effectively utilizing the available relatively few math-related data to improve the performance of the relatively small PLM in complex mathematical problem understanding tasks.
The above situation is common in real-world applications, and is more applicable and affordable than collecting and using large-scale math-related data.
}

\ignore{
To achieve it, we aim to improve the understanding of mathematical knowledge points via the mixture-of-expert multi-task learning, and also enhance the mathematical reasoning capacity by in-context learning using a large language model.
For the understanding of mathematical knowledge points, as the data resource is limited, we consider to make full use of the precise annotated information from the available math-related tasks.
In fact, the examples from different math-related tasks (\eg multi-choice and blank-filling questions) may involve the same knowledge point (\eg the Pythagorean theorem), but are just different in the data format.
Thus, it is promising to integrate various math-related tasks for jointly learning the latent mathematical knowledge points across different instances in these tasks.
Consider that the instances across these tasks require various mathematical knowledge, we also add the Mixture-of-Experts~(MoE) modules~\cite{} into the PLM, to decompose and unify the different mathematical knowledge among all the instances into multiple experts.
Then, the dynamic routing mechanism within the MoE module would select the most relevant experts for each instance to accomplish the corresponding task.
For the mathematical reasoning logic, we introduce a large language model~(LLM) to refine the generated computation and derivation processes from the PLM.
Existing works~\cite{} have revealed that LLMs have learned rich common sense and logic knowledge, and are able to follow the given instruction to accomplish the corresponding task.
Therefore, we devise a retrieval-based iterative prompting strategy to guide the LLM to correct the possible errors in the generated text according to special instructions step by step, without any training.
In this way, we can not only omit the cost of large-scale training, but also fix targeted errors of mathematical reasoning logic in the generated texts. 
}

To this end, in this paper, we propose \textbf{JiuZhang 2.0}, a unified Chinese PLM specially for multi-task mathematical problem solving.  
In order to enhance the multi-task capacity, we make three major technical contributions. 
Firstly, we design a  \emph{Mixture-of-Experts~(MoE)} based architecture to transfer and share mathematical  knowledge across tasks. We adopt the  MoE architecture to encode mathematical text with an elaborately designed routing mechanism.  
Secondly, 
we design \emph{multi-task continual pre-training} and \emph{multi-task fine-tuning} strategies to optimize the MoE-based architecture for multi-task adaptation.  
For multi-task continual pre-training, we construct a group of self-supervised pre-training tasks to warm up the MoE architecture for knowledge sharing; for multi-task fine-tuning, we unify the math-related tasks into two general formats of language  understanding and generation, and
directly enhance the knowledge sharing across these tasks. 
Thirdly, in order to further improve the general capacity of solving different complex tasks, we leverage \emph{LLMs} as complementary models to improve the generated solution by our PLM.  The PLM  (with a smaller tuning cost) is used for task adaptation and generates a preliminary solution, while the LLM  (with a stronger model capacity)  mainly refines the generated results without directly solving the problem. Concretely, we retrieve similar examples and iteratively concatenate instructions with them to compose the prompt,  gradually guiding  the LLM to improve the generation results in a coarse-to-fine manner (overall logic, deduction process and language expressions).

To verify the effectiveness of our proposed JiuZhang 2.0, we conduct extensive experiments on eight tasks, covering both the evaluation settings of \emph{seen tasks} and \emph{unseen tasks}. 
Experimental results have shown that our approach can consistently outperform a number of competitive baseline methods (even LLM based methods). Besides, we deploy our model in a Chinese education app and online $A/B$ test further verifies the effectiveness of our approach.
\section{Related Work}


This work focuses on solving mathematical problems, which has been extensively discussed in the literature~\cite{Sundaram2022WhyAN,Meadows2022ASI,Lu2022ASO}. Various resources or toolkits are released~\cite{lan2021mwptoolkit,cobbe2021training,Hendrycks2021MeasuringMP},  and also empower a variety of math-related applications~\cite{zanibbi2012recognition,zhong2021pya0,banerjee2023deep}. 
In the following, we will review the related study in three major technical approaches.

\paratitle{Traditional NLP Approaches.} 
Since mathematical problems are described in natural language, it is straightforward  to cast the understanding of mathematical problems as a natural language processing~(NLP) task. 
A major difficulty lies in the understanding of the formulas and logic that   mathematical text  contains. Thus, early NLP approaches typically extract the features for understanding the text and formulas, \eg semantic parser~\cite{shi2015automatically} and operator tree~\cite{zanibbi2012recognition}.
In recent years, a surge of methods introduce the deep neural network into mathematical problem understanding. 
They generally leverage advanced NLP models,  \eg RNN~\cite{chiang2019semantically} and Transformer~\cite{li2019modeling},  to encode the mathematical text into meaningful representations.

\paratitle{PLM Based Approaches}. Inspired by the success of PLMs in NLP tasks, researchers employ PLMs to deal with mathematical problems~\cite{peng2021mathbert,polu2020generative}, showing the superiority in understanding and modeling of mathematical texts.
Basically, these methods continually pre-train PLMs (\eg BERT~\cite{devlin2018bert}) with a specific math corpus, and design proper pre-training strategies to capture the semantics of the formulas and logics conveyed in the mathematical texts, \eg text-formula representation alignment~\cite{peng2021mathbert,gong-etal-2022-continual}, basic-to-advanced curriculum pre-training~\cite{zhao2022jiuzhang} and unified multi-task learning~\cite{Mishra2022LilaAU}. 
However,  existing PLM approaches cannot well solve complex mathematical problems and also have a high cost in multi-task deployment. 
 


\ignore{
\paratitle{Chinese Pre-trained Language Models.}
Pre-trained Language Models~(PLMs) (\eg BERT~\cite{devlin-etal-2019-bert}, BART~\cite{lewis-etal-2020-bart} and T5~\cite{raffel2020exploring}) have largely advanced the progress of language intelligence.
Following this direction, our work is based on Chinese PLMs. 
The first line of works adapt BERT~\cite{devlin-etal-2019-bert} by reusing masked language model~(MLM) task to pre-train Transformer encoders~\cite{cui2021pre,sun2021chinesebert} on Chinese corpus. They consider modeling the linguistic characteristics or semantic knowledge of Chinese texts and devise special strategies to improve the task performance, \eg whole word masking~\cite{cui2021pre}, glyph and pinyin embedding~\cite{sun2021chinesebert} and entity enhanced embedding~\cite{jia-etal-2020-entity}. Focused on natural language understanding, these methods can't be directly applied to text generation tasks.
Another line of works pre-train Transformer via the auto-regressive task~\cite{zhang2021cpm,zeng2021pangu} or the seq2seq task~\cite{zhang2021mengzi}. They predict or recover the corrupted tokens  from left to right.
Furthermore,  several studies attempt to endow Chinese PLMs with both capacities of understanding and generation. For example,
Mengzi~\cite{zhang2021mengzi} and ERNIE-3.0~\cite{DBLP:journals/corr/abs-2107-02137}  consist of shared encoding layers and multiple task-specific decoders.
With a similar architecture, CPT~\cite{shao2021cpt} adopts a deeper encoder and two shallower decoders, to accelerate the inference of text generation.  Our work presents the first Chinese PLM for understanding mathematical texts, with a series of specially designed pre-training tasks. }

\paratitle{LLM Based Approaches.}
In contrast to PLMs with moderate sizes, large language models~(LLMs)~\cite{brown2020language, chen2021evaluating, LLMSurvey} are introduced to solve mathematical problems~\cite{Lewkowycz2022SolvingQR,Mishra2022LilaAU,hendrycks2021measuring,cobbe2021training}.   Further,  external modules or tools are used to assist LLMs in complex math problem solving, \eg program interpreter~\cite{drori2021neural,chen2022program,gao2022pal}.
Since it is very costly to tune LLMs, in-context learning~\cite{brown2020language} has been widely used to solve different tasks, \eg chain-of-thought~(CoT) method that uses multi-step  reasoning~\cite{Wei2022ChainOT}. 
Based on CoT, several improvements have been proposed for mathematical reasoning, including  selecting more appropriate samples~\cite{Fu2022ComplexityBasedPF,Zhang2022AutomaticCO}, designing better instructions~\cite{Kojima2022LargeLM}, generating multiple results for ranking~\cite{wang2022self,li2022advance,Zhu2022SolvingMW} and decomposing problem into sub-problems~\cite{zhou2022least}.
However, it is hard for LLMs to adapt to the domains or tasks with large differences from the pre-training setting~\cite{hendrycks2021measuring}, \eg Chinese mathematical problem solving.  


\ignore{
For math-related tasks, several works also collect a very large number of math-related texts for pre-training LLMs~\cite{Lewkowycz2022SolvingQR,Mishra2022LilaAU}, which greatly improves their fine-tuning performance.
Based on these LLMs, recent works further integrate external modules or toolkits to assist LLMs in complex math problem solving, \eg program interpreter~\cite{drori2021neural,chen2022program,gao2022pal}.
Despite the remarkable performance, it is very costly to fine-tune these LLMs.
Therefore, a typical way called in-context learning~\cite{brown2020language} has been proposed, where the LLM is not tuned and just directly generates the results based on sampled few examples and instructions in the input.
Based on in-context learning, the chain-of-thought method~(CoT)~\cite{Wei2022ChainOT} has been proposed, which adds detailed description about intermediate reasoning steps into the input, guiding LLMs to think step-by-step until generating the answer.
Due to the huge improvement of CoT on multi-step mathematical reasoning tasks, several works attempt to optimize this way, \eg selecting more appropriate samples~\cite{Fu2022ComplexityBasedPF,Zhang2022AutomaticCO}, designing better instructions~\cite{Kojima2022LargeLM}, generating multiple results for ranking~\cite{wang2022self,li2022advance}, decomposing problem into sub-problems~\cite{zhou2022least}.
Despite the success of the above methods, as in-context learning methods do not tune the parameters, they mainly rely on the pre-learned knowledge of LLMs and are hard to utilize specific mathematical knowledge beyond the pre-training corpus~\cite{cobbe2021training,hendrycks2021measuring}.
}

Besides, our model is built on MoE architecture~\cite{Jacobs1991AdaptiveMO}, which aims to scale up the model capacity with controllable  computational cost. For MoE architectures, it is important to design suitable expert  
network~\cite{Ponti2022CombiningMS}, routing mechanism~\cite{Kudugunta2021BeyondDT,Ye2022ElicitingAU,Gupta2022SparselyAM} and training strategies~\cite{Ye2022ElicitingAU,Shazeer2017OutrageouslyLN,Zoph2022STMoEDS}. While, our work has presented a novel application of MoE for dealing with mathematical tasks, with specific improvements.   
Our work is also related to multi-task learning based on language models~\cite{Liu2019MultiTaskDN,Aghajanyan2021MuppetMM}, while our focus is to share mathematical knowledge across. We design specific architecture and corresponding training strategies for mathematical problem solving, which distinguishes it from prior work on multi-task learning.

\ignore{
\paratitle{Mixture-of-Experts.}
The mixture-of-experts~(MoE) layer~\cite{Jacobs1991AdaptiveMO} has been widely used in deep neural networks, which can scale up the model capacity and only introduce a small computation cost.
Generally, an MoE layer contains several expert networks that share the same network structure but own different parameters, and a sparse routing function~(or network)~\cite{Shazeer2017OutrageouslyLN} will be utilized to select a few experts at once for each input.
However, there are also several common issues in training and using the MoE layer, \eg the unbalanced load among all expert candidates~\cite{Shazeer2017OutrageouslyLN} and the training instability problem~\cite{Zoph2022STMoEDS}.
To address them, several works propose special regularization strategies to encourage the training stability and uniform use of the MoE layer, and further improve the performance.
Recently, the MoE layer has also been introduced into multi-task learning for balancing the ability of general and task-specific skills~\cite{Kudugunta2021BeyondDT,Gupta2022SparselyAM,Ye2022ElicitingAU}.
With the help of the learnable routing network, the MoE layer can adaptively decompose and preserve the knowledge required for different tasks, avoiding interference among the conflict task-specific knowledge.
Following this way, a line of works further optimize the selection of experts, by devising the token-level routing mechanism~\cite{Kudugunta2021BeyondDT,Ye2022ElicitingAU}, incorporating additional task embeddings~\cite{Gupta2022SparselyAM} and adopting multi-stage training~\cite{Ye2022ElicitingAU}. 
}

\ignore{\section{Preliminary}
\subsection{Problem Statement}
We first formally describe  \emph{mathematical text},  a general phrase referring  to the text relating to a mathematical problem. 
In our corpus,  a mathematical problem is associated with two texts, namely problem statement and solution text (\aka answer key). 
The problem statement introduces necessary background information and explains the problem that is to be solved, and  the answer key describes the key hints or complete solving  procedure for deriving the answer. 
We concatenate both  problem statement and solution description as the \emph{mathematical text} of a mathematical problem. 
Overall, a mathematical text is a  textual description that mixes text words with math symbols. Given a mathematical problem $q$, the corresponding mathematical text can be considered as a sequence of $n$ tokens, denoted as $q=\{t_1, t_2, \cdots , t_n\}$, where each token $t_i$ is either a text word or a math symbol (denoting a variable or an operator).  Furthermore, a consecutive segment of $l$ math symbols constitute a math formula, denoted as $f_i = \{s_1, \cdots, s_l\}$, where each math symbol $s_j$ is from $\{t_1, t_2, \cdots , t_n\}$. There are usually multiple formulas in a mathematical text, denoted as $\{f_1, f_2, \cdots , f_m\}$.
Based on the above notations, this work focuses on pre-training a PLM on a large-scale corpus consisting of  
mathematical texts. Then, the PLM can be 
fine-tuned on various mathematical tasks (\eg knowledge point classification and similar question retrieval), and improve the corresponding task performance.
}

\section{Approach}
In this section, we present our \textbf{JiuZhang 2.0},  which is developed based on the former version of JiuZhang by introducing specific improvements for multi-task mathematical problem solving. 

\begin{figure*}
  \centering
  \includegraphics[scale=0.6]{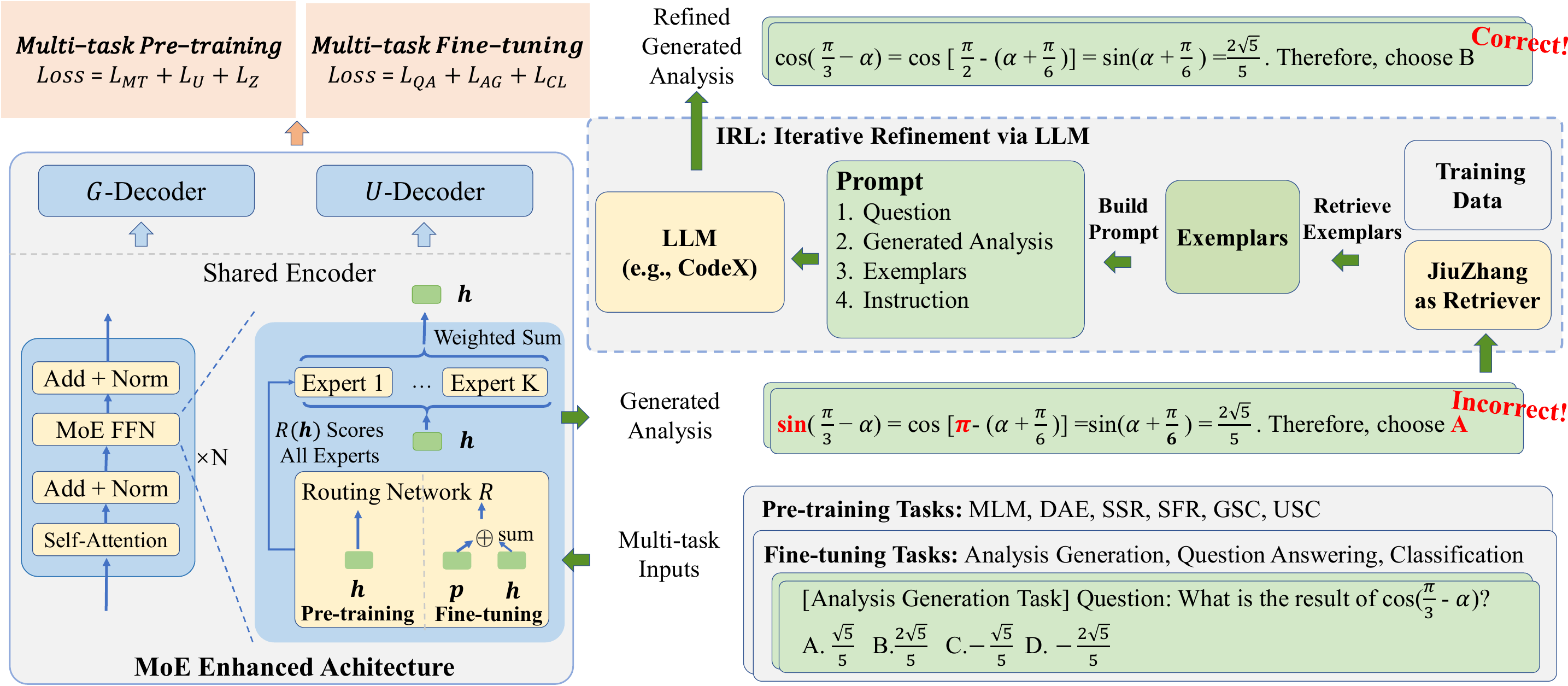}
  \caption{The overview of our model JiuZhang 2.0, consisting of two major parts: MoE extension with multi-task training based on the PLM (the primary role) and iterative refinement via LLM (the complementary role). {The red bold tokens are errors generated by JiuZhang, which are corrected by LLM in the later iterative refinement process.}}
  \label{approach}
\end{figure*}

\subsection{Backbone Model: JiuZhang}
\label{jz_intro}

We first introduce the backbone model JiuZhang~\cite{zhao2022jiuzhang} for mathematical problem understanding. 
Unlike general-purpose PLMs (\eg BERT~\cite{devlin2018bert}), JiuZhang considers the pre-training corpus of \emph{mathematical text}, in which each  text consists of  a sequence of $n$ tokens (either a text word or a math symbol) corresponding to a mathematical problem (including both problem statement and possible solution), denoted as $d=\{t_1, t_2, \cdots , t_n\}$.  Next, we introduce the original architecture and pre-training tasks for JiuZhang~\cite{zhao2022jiuzhang}.





\paratitle{Architecture}. 
Since  both understanding and generation capacities are needed for mathematical problem solving, JiuZhang adopts an architecture consisting of \emph{one shared encoder} and \emph{two task-specific decoders}: one decoder for understanding tasks ($U$-decoder) and the other decoder for generation tasks ($G$-decoder).  
It employs  bidirectional Transformers to implement the shared encoder and the $U$-decoder, and an auto-regressive Transformer to implement the $G$-decoder. 
In order to enhance the representation ability, the  shared encoder is built with more layers than the  two decoders (\ie 10 layers \emph{v.s.} 2 layers). 
Given a mathematical text $d=\{t_1, \cdots, t_{n}\}$, the shared encoder can 
produce contextualized token representations $\{\mathbf{h}^{(L)}_{1}, \mathbf{h}^{(L)}_{2}, \cdots, \mathbf{h}^{(L)}_{n}\}$ ($L$-layer architecture) by capturing mathematical semantics from the input text. 
Then, the $U$-decoder and $G$-decoder will solve the understanding and generation tasks based on the contextualized representations, respectively. 


\paratitle{Pre-training Tasks}. In the former version,  JiuZhang sets up three types of pre-training tasks and schedules them in a curriculum learning approach. 
The basic course is constructed based on masked token prediction following general-purpose PLMs, with two pre-training tasks of {masked language modeling} ($L_{MLM}$) and {denoised auto-encoder}  ($L_{DAE}$). 
The advanced course is constructed based on specific considerations of mathematical text, 
including mathematical logic recovering and solution checking.   
For mathematical logic recovering, we introduce the pre-training tasks of shuffled sentences recovering   ($L_{SSR}$) and shuffled formulas recovering   ($L_{SFR}$), in order to enhance the understanding of mathematical logic;  
for solution checking, we introduce the pre-training tasks of dual-decoder solution checking  ($L_{GSC}$ and $L_{USC}$), which improve the model's ability to detect and correct errors in its own generated outputs. 
These pre-training tasks can gradually adapt JiuZhang to mathematical problem solving. 
Due to space limit, please refer to original paper~\cite{zhao2022jiuzhang} for more details.

Although JiuZhang can better model  mathematical text compared with general-purpose PLMs, it is not specially designed for multi-task mathematical problem solving. 
In order to enhance the multi-task capacity, we next introduce two important improvements, namely MoE extension with multi-task training (Section~\ref{sec:moe}) and iterative refinement with LLM  (Section~\ref{sec:llm}).
In the following, we introduce the two parts in detail. 

\subsection{MoE Extension with Multi-task Training}
\label{sec:moe}
By leveraging  a corpus of mathematical text, JiuZhang implicitly captures  mathematical  knowledge with specially designed pre-training tasks. 
While, such information is encoded via a whole model (\ie the shared encoder), and it  is difficult to transfer mathematical knowledge across different tasks. 
To better decompose and share the  mathematical knowledge, we propose to enhance  the backbone model with Mixture-of-Experts~(MoE)~\cite{Shazeer2017OutrageouslyLN} extension, and introduce  multi-task continual pre-training and multi-task fine-tuning strategies  based on MoE-enhanced architecture. 



\subsubsection{MoE Extension for Knowledge Sharing} 

 MoE~\cite{Jacobs1991AdaptiveMO} is a widely used technique  to increase model capacity by  incorporating multiple expert networks (the same architecture yet different parameters). While, we employ MoE to decouple and share mathematical knowledge across tasks: common knowledge for related tasks can be captured  in one specific expert and less irrelevant knowledge across different tasks is distributed among multiple  experts.  




\paratitle{MoE Layer for Mathematical Text.} 
In our approach, we only extend the deep shared encoder (capturing the \emph{essential} mathematical knowledge) with MoE, but not the shallow decoders (supporting different types of tasks).   
As the encoder is composed of multiple bidirectional Transformer layers, we incorporate the MoE layer to substitute for the original feed-forward layer.
Each MoE layer consists of a routing network $R(\cdot)$ and multiple expert networks $\{E_{i}(\cdot)\}_{i=1}^{K}$, where $K$ denotes the number of expert candidates. 
To reuse the encoded knowledge from JiuZhang, we utilize the parameters of its  feed-forward layer to initialize the parameters of the expert networks, which can also improve the training stability. Since a mathematical problem is usually related to diverse knowledge points, we adopt a token-wise routing mechanism~\cite{Shazeer2017OutrageouslyLN} to decouple its associated mathematical knowledge, by assigning experts individually for each token. 
Given an input mathematical text $d=\{t_1, \cdots, t_n\}$, in each Transformer layer, the multi-head self-attention layer  first produces the aggregated representations of all these tokens $\{\bm{h}_1, \cdots, \bm{h}_n\}$.
Then, for each token, the routing network  estimates the  probability distribution over the $K$ experts: 
\begin{align}
    R(\bm{h}) = \text{softmax}\big(\bm{W} \cdot \bm{h} \big), 
    \label{eq-gating-distribution}
\end{align}
where $\bm{W}$ is the trainable matrix for deriving the routing distribution. 
Further, we  employ a weighted combination to integrate the outputs from the $K$ experts: 
\begin{align}
    \text{MoE}(\bm{h}) = \sum_{i=1}^{K}{R(\bm{h})_i \times E_{i}(\bm{h})}. 
    \label{eq-gating-network}
\end{align}

\paratitle{Sparsely Routing with Jitter Noise.}
To save the computational cost in MoE layers, we introduce the sparse activation mechanism~\cite{Shazeer2017OutrageouslyLN} to selectively utilize expert networks for each token. Specifically, according to the estimated probability distribution $R(\bm{h})$, we first rank all the expert networks and then select the top-$k$ ones ($k \leq K$) in Eq.~\eqref{eq-gating-network}  to derive the token representation.
Here, we set $k=1$, \ie only the most related expert will be routed for each token. 
In this way, for each token, the computational cost of the expert network is roughly the same as the original feed-forward layer of JiuZhang. More detailed analysis about  \emph{inference latency} can be found in Appendix~\ref{sec:latency}. 
However, prior studies~\cite{fedus2021switch} have found that such a sparse expert assignment approach would deterministically choose the best-ranking expert, causing the expert network easy to overfit.
Therefore, we introduce randomness into the expert selection process by using the \emph{jitter noise}~\cite{fedus2021switch} in the routing network.
We multiply the estimated probability distribution in Eq.~\eqref{eq-gating-distribution}
by a jitter noise $\bm{\epsilon}$ (a randomly scaling distribution vector) as:
\begin{align}
    R(\bm{h}) = \text{softmax}\big( (\bm{W} \cdot \bm{h}) \odot \bm{\epsilon}\big), 
    \label{eq-jitter-noise}
\end{align}
where $ \bm{\epsilon} \in \mathbb{R}^K$ is a randomly sampled vector and each entry is from a uniform distribution $[1-\eta, 1+\eta]$  (with the noise degree controlling hyper-parameter $\eta$), and  ``$\odot$'' is the element-wise product. 
In this way, the probability scores of different experts would be increased or decreased randomly, making the expert networks more robust to perturbations on the routing results.

\subsubsection{ Multi-task Pre-training for MoE Adaptation}
In order to support the MoE architecture, we  design multi-task continual pre-training strategies for adapting to the multi-task setting. 

\paratitle{Multi-task Continual Pre-training.}
The goal of multi-task pre-training is to decouple and transfer mathematical knowledge via expert sharing, according to task  supervision. Since there is no task data during the pre-training stage, we consider reusing the original pre-training tasks of  JiuZhang discussed in Section~\ref{jz_intro}, including 
masked token prediction ($L_{MLM}$ and $L_{DAE}$), mathematical logic recovering ($L_{SSR}$ and $L_{SFR}$) and solution checking ($L_{GSC}$ and $L_{USC}$). 
Instead of using a  curriculum learning way as in \cite{zhao2022jiuzhang}, we treat the six pre-training losses as equal optimization goals, and set a multi-task pre-training objective:
\begin{align}
    L_{MT} = L_{MLM} + L_{DAE} + L_{SSR} + L_{SFR} + L_{USC} + L_{GSC}. 
    \label{eq-continue-pretrain}
\end{align} 
Note that our model has been initialized with the parameters of  the former JiuZhang, so that it also implicitly benefits from the curriculum learning strategy proposed in the previous paper~\cite{zhao2022jiuzhang}.   
While, based on the MoE-based architecture, we employ these pre-training tasks to decouple and share mathematical knowledge across tasks.  


\ignore{As mentioned in Section~\ref{jz_intro}, Jiuzhang utilizes the curriculum pre-training strategy that progressively learns three types of pre-training tasks, corresponding to different mathematical abilities from basic to advanced.
In fact, since these pre-training tasks are based on the same pre-training corpus, for the same instance, they may also share the same mathematical knowledge points to predict the masked tokens, recover shuffled formulas and correct the generated solutions.
Therefore, we directly integrate them in a multi-tasking manner, and the training objective is
\begin{align}
    L_{MT} = L_{MLM} + L_{DAE} + L_{SSR} + L_{SFR} + L_{USC} + L_{GSC},
    \label{eq-continue-pretrain}
\end{align}
where the input and output data formats are the same as in Section~\ref{}.
}

\paratitle{Auxiliary Losses for Improved Optimization.}  
For MoE methods, there are two major training problems that affect the performance, 
\ie the unbalanced load among experts~\cite{Shazeer2017OutrageouslyLN} and the training instability~\cite{Zoph2022STMoEDS}. 
To alleviate these problems, we adopt two auxiliary losses~\cite{Shazeer2017OutrageouslyLN,Zoph2022STMoEDS}  as the regularizers in our approach. 
Specially, the unbalanced load problem refers that certain experts are extremely frequently routed, which may cause the overfitting problem on these experts and the underfitting problem on other experts.
Therefore, we aim to improve the unbalanced routing among all $K$ experts.
Formally, we encourage the accumulated estimated probabilities for each expert to be uniform, denoted as:
\begin{align}
    L_{U} = \alpha \cdot K \cdot \sum_{i=1}^{K}{f_i \cdot s_i},
    \label{eq-auxiliary-1}
\end{align}
where $f_i$ is the number of tokens dispatched to the  $i$-th expert, and $s_i$ is the accumulated routing score estimated by the routing network for the  $i$-th expert, and $\alpha$ is the coefficient to control the influence.
According to \cite{Shazeer2017OutrageouslyLN}, this loss encourages uniform routing since it would be minimized under a uniform distribution. 
Further, the training instability problem is often caused by the large volatility of the probability  scores in the routing network. 
In order to control the  volatility, we adopt the $Z$-loss~\cite{Zoph2022STMoEDS} that encourages the routing logits of all tokens (size $n$) to remain small as:
\begin{align}
    L_Z = \beta \cdot \frac{1}{n} \log \sum_{j=1}^{n} \exp\big(R(\bm{h}_j)\big)^2,
    \label{eq-auxiliary-3}
\end{align}
where $\beta$ is the coefficient for this loss.



\subsubsection{Multi-task Fine-tuning for MoE Adaptation}
To apply the pre-trained model, a typical way is to fine-tune it on some downstream tasks. While, it cannot sufficiently leverage the merits of MoE-based architectures (\ie decoupling and sharing), without considering inter-task relationships.  Thus, we  design a multi-task fine-tuning strategy, which boosts the capacity of our MoE architecture by leveraging the data of all (available) downstream tasks. 


\paratitle{Unifying the Fine-tuning Tasks.}
For multi-task fine-tuning, we combine the available training data from multiple downstream tasks for jointly optimizing  our model. 
Since these tasks that we consider are math related, they tend to rely on common mathematical knowledge for task solving, which can be captured via the MoE-based architecture. 
However, the formats of the input and output data for downstream tasks are generally different, making it hard to be jointly fine-tuned.
Recall that our backbone model has included two specific decoders that can handle both  understanding and generation tasks for mathematical text. 
Thus, we unify the  math-related tasks into two general formats, either \emph{understanding} or \emph{generation}. 
Specially, for all text classification tasks, we  merge the annotation labels  and consider an extended multi-label setting, where the label dictionary covers  the labels from all classification tasks.  
In this way, we can equip our $U$-decoder with a multi-label classifier head to simultaneously accomplish all these classification tasks. 
Further, for all text generation tasks, we adopt a standard sequence-to-sequence format and  utilize the $G$-decoder to solve them.  
To better distinguish the different tasks for our model, given the training data from $m$ tasks, we also devise $m$ task prompt embeddings, denoted as $\{\bm{p}_1, \cdots, \bm{p}_m\}$.
For each instance, we insert its task prompt embedding after the $\textsc{[CLS]}$ token embedding.

\paratitle{Routing with Task Prompt.}
During multi-task fine-tuning, as the task type may be useful to determine the selection of different experts with specific mathematical knowledge, we further revise the routing mechanism by incorporating  task-level instruction.
Specially, in each MoE layer, we add the input token representation $\bm{h}$ with the representation of the task prompt $\bm{p}$, to compose the input of the routing layer for estimating the probability distribution over the experts as:
\begin{equation}
    R(\bm{h}) = \text{softmax}\big((\bm{W} \cdot (\bm{h}+\bm{p}) ) \odot \bm\epsilon \big), 
    \label{eq-ft-routing}
\end{equation}
where we also use  jitter noise to improve the robustness. 



\subsection{Iterative Refinement via LLM}
\label{sec:llm}
Although  MoE extension is employed to enhance the backbone model, we keep a moderate-sized model (\ie $276M$ for $K=4$) with an affordable cost for downstream applications.  Due to the limit in model size and pre-training data, it still has difficulty in generating solution text for some complex mathematical problems. 
Our solution is to leverage large language model (LLM)~\cite{brown2020language,chen2021evaluating} with stronger general modeling capacities for refining the generation results of our PLM. 
\ignore{
Actually, due to the limitation of pre-training data, existing PLMs are easy to make mistakes during generation, \eg calculation error and transcription error.
Such a problem can be solved by large language models, since they have been pre-trained on large-scale general corpus, owning better capacity of generating fluent and logically correct texts.
To effectively leverage the LLM, we propose to utilize the in-context learning method~\cite{brown2020language} to guide the LLM to iteratively refine the generated results of our model.
To achieve it, the key is to compose high-quality prompts as the input to instruct the LLM, which generally consists of a clear instruction and a few sampled helpful examples as references.
}
To achieve this, we first design a retrieval strategy to select the most relevant exemplars for constructing the prompts, and then devise an iterative prompting method that utilizes in-context learning to gradually correct the generated results.


\subsubsection{Constructing Prompts Using Retrieved Samples}
Since existing LLMs are mainly English-focused, they cannot sufficiently capture  the necessary mathematical knowledge to effectively accomplish math-related tasks in Chinese (see experiments in Section~\ref{sec:main-exp}). Thus, instead of directly solving the tasks,  LLM plays \emph{a complementary role} in our approach for  refining the generated results of our PLM. 
Specifically, given a mathematical problem $q$, we first utilize the PLM (Section~\ref{sec:moe}) to generate the solution text $\hat{a}$, and then employ the LLM via in-context learning~\cite{brown2020language}  to refine $\hat{a}$ into $\tilde{a}$ with improved quality.  
To provide effective guidance on the LLM, we construct the prompts with retrieved relevant exemplars and specially designed  natural language instructions.

\paratitle{Retrieving Exemplars.}
As empirical studies~\cite{Liu2021WhatMG} have revealed that the exemplars in the prompts of LLMs are important to the task performance, we retrieve relevant instances from the training data as the exemplars. Since exemplar finding is essentially an unsupervised text retrieval task, we further employ SimCSE~\cite{Gao2021SimCSESC} to enhance the representation capacity of our backbone model for semantic matching. Following  SimCSE, we incorporate the dropout mechanism to augment positive representations and utilize the contrastive learning objective for training.  
In the retrieval stage, given the target problem $q$ and the training data set as the retrieval candidate pool, we first encode all the mathematical problems into dense vectors by our backbone model, and then select the top-ranking problems as relevant exemplars, denoted as $C = \{ \langle q_j, a_j \rangle \}^B_{j = 1}$, where $a_j$ is the associated solution text for problem $q_j$.  Note that we do not use the solution text for the target problem, while only utilizing the solution texts of the problems from  training data. 




\paratitle{Building Prompts.}
In order to guide the LLM to refer to the retrieved exemplars for revising the generated result $\hat{a}$ from our PLM, we utilize the in-context learning method with specially designed prompts. Specifically, the input of the LLM consists of four parts, \ie the given question $q$, the generated result $\hat{a}$, the retrieved exemplars $C = \{ \langle q_j, a_j \rangle \}^B_{j = 1}$, and a natural language instruction $I$.
We concatenate the above four parts  into a long sentence, to compose the prompt template as:
\begin{equation}\label{eq-prompt}
    [q; \hat{a}; C; I] \rightarrow \text{prompt(LLM)}, 
\end{equation}
where the instruction $I$ can be flexibly set according to different tasks. We will discuss how to set it in the following part. 


\subsubsection{Iterative Prompting for Result Refinement}
Generally, the generated results from the PLM may contain a variety of mistakes (\eg inconsistent logic and language typos),  and  it is hard for the LLM to completely check and correct all these mistakes at once. 
Therefore, we devise a three-stage iterative refining strategy that gradually improves the generated results following a coarse-to-fine manner.
Concretely, based on the prompt template in Eq.~\eqref{eq-prompt},  we design three specific instructions for the three stages, which guide the LLM to refine the generation results from the three perspectives of  \emph{overall logic}, \emph{deduction process} and \emph{language expressions}, respectively. We present the above instructions  in the Appendix~(Table~\ref{iter-setting}).   

Further, to better cooperate with the above instructions, we also revise the way of retrieving exemplars in the three stages: 
\begin{itemize}
\item at the first stage, we only rely on the problem statement $q$ for finding similar problems, referring  to their overall logic; 
\item at the second stage, we leverage both $q$ and the generated solution text $\hat{a}$ for retrieving relevant problems with similar solution text,  checking the deduction process; 
\item at the third stage, we only utilize the generated solution text $\hat{a}$ for retrieval to find other similar solution texts,   correcting improper language expressions. 
\end{itemize}

To accomplish the goal for each individual stage, we find that it needs multiple iterations for LLM to produce ideal outputs. 
Thus, we perform $T$-step ($T=3$) iterations for each stage. 
At each step, the  refined output $\Tilde{a}^{(t)}$ will be used as the input of 
the next step $\hat{a}^{(t+1)}$ to compose the prompt and the retrieved exemplars can also be updated according to new query $\hat{a}^{(t+1)}$. 
In this way, we can iteratively refine the generated results until the expected  goal is fulfilled  at each stage, and finally generate high-quality results.

\ignore{
The process of iteratively answer correction can be divided into three stages: Idea Correction Stage, Error Correction Stage and Language Expression Correction Stage. 
Each stage has unique retrieval methods and prompt templates and will repeat several times.
Formally, we design a set of retrieval methods and prompt templates, denoted as $R=\{r_1, r_2, r_3\}$ and $P=\{p_1, p_2, p_3\}$ respectively. 
The $t$-th stage uses $r_t$ as a retrieval method and uses template $p_t$ to construct a demonstration to prompt LLM. 
Each stage has a hyper-parameter $k$, which means the iteration steps of the corresponding stage.
The three stages of answer correction instruct LLM to revise JiuZhang's solution step by step ranging from coarse to fine-grained. The details of retrieval methods and instructions in each stage prompt are shown in Table~\ref{iter-setting}.

\paratitle{Idea Correction Stage}
For each mathematical problem $q$, the problem content $c$ is used to retrieve similar problems in this stage. 
The problem content $c$ will be encoded by the word2vector model to a vector $v_c \in \mathbb{R}^d$ as the embedding of the problem to retrieve similar problems from the retrieval set.
It's empirical that little difference in the content will lead to differences in the ideas for solving problems most of the time.
Using problem content as key to retrieval can retrieve similar problem descriptions and diversified solutions from the retrieval set.
The corresponding instruction in the prompt guides LLM to select correct idea for the problem.

\paratitle{Error Correction Stage}
In this stage, we use problem content $c$ and analysis (proper analysis for retrieval set problems while generated analysis for test set problems) to retrieve examplars.
The vector $v_c \in \mathbb{R}^d$ and $v_a \in \mathbb{R}^d$ are represented the embedding of problem content and analysis respectively.
The final embedding of a problem $v \in \mathbb{R}^{2d}$ used in retrieval is the concatenating of content embedding $v_c$ and analysis embedding $v_a$.
The concatenating of $v_c$ and $v_a$ forces the retriever to retrieve problems which have similar problem content and analysis with the test problem.
During this stage, we provide more related problems both in problem description and idea of solution as examplars to LLM.
This stage helps LLM to correct the calculation errors and logic errors in the generated analysis from the last iteration.

\paratitle{Language Expression Correction Stage}
Given a mathematical problem of test set, its analysis is generated by JiuZhang and corrected by LLM through the first two stages.
So the target of the third stage is to improve the fluency of generated analysis and correct the remaining errors in it.
We encode problem analysis (generated analysis for problem in test set) to an embedding $v_a \in \mathbb{R}^d$ and use it to retrieve the examplars.
The examplars provided to LLM in this stage have high similarity in analysis with problems in the test set. 
The instruction in this stage guides LLM to focus on the difference in language expression between reference and proper analysis.
This stage helps LLM to perfect the solution and fix the remaining minor mistakes. }
\section{Experiments}

\subsection{Experimental Settings}
We utilize the same pre-training corpus of JiuZhang~\cite{zhao2022jiuzhang}, consisting of 1,276,952 high-school math problems collected from Zhixuewang, and each problem is associated with the problem type, problem statement and solution text. 
We preprocess these collected texts in the same way as JiuZhang.

\begin{table}[t]
    \centering
    \caption{Statistics of the datasets for eight evaluation tasks. ``Seen'' and ``Unseen''  refer that the task data is \emph{used} or \emph{not used} during multi-task fine-tuning, respectively.}
    \begin{tabular}{ccrrrr}
        \bottomrule
        \textbf{Setting} & \textbf{Type} & \textbf{Task} & \textbf{Train} & \textbf{Dev} & \textbf{Test} \\ 
        \hline
        \multirow{6}*{Seen} & \multirow{2}*{QA tasks} & MCQ & 22,000 & 3,982 & 7,466 \\
            & & BFQ & 14,795 & 1,786 & 1,778 \\
            \cline{2-6}
            & \multirow{2}*{Generation} & CAG & 16,000 & 1,976 & 1,977 \\
            & & BAG & 14,795 & 1,786 & 1,778 \\
            \cline{2-6}
            & \multirow{2}*{Classification} & KPC & 8,721 & 991 & 1,985 \\
            & & QRC & 10,000 & 2,000 & 4,000 \\
           
        \hline
        \multirow{2}*{Unseen}
            & \multirow{2}*{Generation} & JCAG & 8,000 & 1,000 & 1,000 \\
            & & JBAG & 8,000 & 1,000 & 1,000 \\
        \bottomrule
    \end{tabular}
    \label{tab-static}
\end{table}

\paratitle{Evaluation Tasks.} We consider two different settings for evaluation, namely \emph{seen tasks} and \emph{unseen tasks},  referring to the task data that are \emph{used} and \emph{not used}, respectively,   during multi-task fine-tuning. We split each task dataset into training/development/test sets. The statistics of these tasks are shown in Table~\ref{tab-static}.

$\bullet$ \emph{Seen tasks} consist of six tasks based on high-school math problems, including (1) two question answering tasks, \ie Multiple-Choice Question Answering (MCQ) and Blank-Filling Question Answering (BFQ); (2) two analysis generation tasks, \ie 
        Multiple-Choice Analysis Generation (CAG) and Blank-Filling Analysis Generation (BAG); and (3) two classification tasks, \ie Knowledge Point Classification (KPC) and Question Relation Classification (QRC). 
 For these tasks, we perform multi-task fine-tuning with all training sets, select the model  checkpoint with the best average performance on development sets, and then evaluate the results on test sets.

$\bullet$ \emph{Unseen tasks} consist of two analysis generation tasks based on junior high school math problems, \ie Junior-high-school Multiple-Choice Analysis Generation (JCAG) and Junior-high-school Blank-Filling Analysis Generation (JBAG), which are \emph{not used} in  multi-task fine-tuning for our model. For the two tasks, we perform task-specific fine-tuning, \ie  the multi-task fine-tuned model is separately optimized, tuned and evaluated for each task. 



We use the evaluation metrics following JiuZhang~\cite{zhao2022jiuzhang}.
For classification tasks (KPC and QRC), we adopt Accuracy and F1-macro as the evaluation metrics.
For question answering tasks (MCQ and BFQ), we adopt Accuracy for evaluation. 
For generation tasks (CAG, BAG, JCAG and JBAG), we use BLEU-4~\cite{Papineni2002BleuAM}, ROUGE-2 and ROUGE-L~\cite{Lin2004ROUGEAP} to evaluate the quality of the generated analysis, and also adopt Accuracy to evaluate the generated answers.

\paratitle{Baseline Methods.}
We select the following four types of baselines: 

$\bullet$ \emph{Non-pretraining methods} consist of classic neural network methods for text classification or generation, \ie TextCNN~\cite{kim-2014-convolutional}, TextRCNN~\cite{lai2015recurrent}, Seq2Seq~\cite{Bahdanau2015NeuralMT} and Transformer~\cite{vaswani2017attention}.

$\bullet$ \emph{Pre-trained language models} have been pre-trained on large-scale general corpus. We select BERT-Base~\cite{devlin-etal-2019-bert}, BART-Base~\cite{lewis-etal-2020-bart}, RoBERTa-wwm~\cite{cui2021pre}, CPT~\cite{shao2021cpt} and Mengzi~\cite{zhang2021mengzi}. For generation tasks,  we fine-tune RoBERTa-wwm in a UniLM way~\cite{Dong2019UnifiedLM}, and utilize bi-directional attention for input and unidirectional attention for output to implement the Seq2Seq based  training and inference. 

$\bullet$ \emph{Continual pre-training methods} further pre-train PLMs on domain-specific corpus (our collected math corpus), and also adopt specially designed pre-training tasks.
We select MathBERT~\cite{peng2021mathbert}, DAPT-BERT~\cite{gururangan-etal-2020-dont}, DAPT-CPT, COMUS~\cite{gong-etal-2022-continual}, JiuZhang~\cite{zhao2022jiuzhang}.
Since our approach is also related to multi-task learning~\cite{Liu2019MultiTaskDN,Aghajanyan2021MuppetMM}, we also add a variant that extends  JiuZhang~\cite{zhao2022jiuzhang} in a multi-task training strategy, MTDNN~\cite{Liu2019MultiTaskDN} for fine-tuning. 

$\bullet$ \emph{Chain-of-thought~(CoT) methods} {add explanations to the exemplars in the input prompt of LLMs}, to better guide them to generate correct answer~\cite{Wei2022ChainOT}.
We employ CoT on GPT-3~\cite{brown2020language} and CodeX~\cite{chen2021evaluating}, \ie GPT3-CoT and CodeX-CoT.

\begin{table*}
\centering
\caption[Caption for LOF]{Main results on two question answering tasks and two analysis generation tasks in the setting of seen tasks. Here, ``Acc.'' denotes the metric Accuracy, and ``w/o IRL'' denotes removing the iterative refinement strategy using LLMs.
The best and the second-best methods are denoted in bold and underlined fonts respectively. }
\begin{tabular}{lcccccccccc}
\bottomrule
\textbf{Tasks} & \textbf{MCQ} & \textbf{BFQ} & \multicolumn{4}{c}{\textbf{CAG}} & \multicolumn{4}{c}{\textbf{BAG}} \\
\cmidrule(r){1-1}\cmidrule(r){2-2}\cmidrule(r){3-3}\cmidrule(r){4-7}\cmidrule(r){8-11}
Metrics & Acc. & Acc. & BLEU-4 & ROUGE-2 & ROUGE-L & Acc. & BLEU-4 & ROUGE-2 & ROUGE-L & Acc. \\
\hline
Seq2Seq & 37.61 & 44.32 & 39.91 & 47.79 & 67.88 & 42.63 & 39.86 & 48.15 & 68.06 & 39.91 \\
Transformer & 35.33 & 46.57 & 41.39 & 48.50 & 67.09 & 41.02 & 41.91 & 48.80 & 67.76 & 45.95\\
\hline 
RoBERTa-wwm & 37.29 & 47.24 & 47.29 & 53.81 & 70.61 & 47.70 & 44.62 & 51.5 & 69.54 & 42.35 \\
BART & 36.15 & 46.82 & 48.20 & 55.04 & 71.66 & 48.92 & 45.46 & 52.16 & 69.62 & 43.92 \\
CPT & 37.90 & 46.31 & 47.98 & 54.97 & 71.67 & 47.03 & 44.82 & 52.29 & 70.01 & 40.68 \\
DAPT-CPT & 46.26 & 53.41 & 49.54 & 55.97 & 72.52 & 50.46 & 46.33 & 53.69 & 70.91 & 48.98 \\
JiuZhang & 47.73 & 54.60 & 50.05 & 56.51 & 72.99 & 54.51 & 47.73 & 54.36 & 71.17 & 51.82\\
JiuZhang-MTDNN & 48.81 & 54.95 & 49.15 & 56.28 & 72.77 & 56.80 & 47.58 & 54.16 & 71.22 & 53.09 \\
\hline
GPT3-CoT & 36.15 & 50.39 &  46.93 & 53.59 & 70.65 & 55.18 & 45.82 & 52.35 & 69.43 & 50.39 \\
CodeX-CoT & 40.36 & 53.82 & 43.65 & 54.28 & 70.43 & 56.30 & 42.96 & 53.45 & 69.89 & 53.82 \\
\hline
JiuZhang 2.0 w/o IRL & \underline{49.75} & \underline{55.85} & \underline{50.17} & \underline{56.72} & \underline{73.02} & \underline{58.83} & \underline{48.33} & \underline{54.79} & \underline{71.48} & \underline{54.78}\\
JiuZhang 2.0 & \textbf{50.37} & \textbf{58.77} & \textbf{50.72} & \textbf{56.97} & \textbf{73.14} & \textbf{60.19} & \textbf{49.39} & \textbf{55.61} & \textbf{71.69} & \textbf{58.77} \\

\bottomrule
\end{tabular}
\label{tab-hard-results}
\end{table*}

\begin{table}
\centering
\caption{Main results on two basic classification tasks in the seen setting. Iterative refinement via LLM is not applicable to the two tasks.  }
\begin{tabular}{lcccc}
\bottomrule
\textbf{Tasks} & \multicolumn{2}{c}{\textbf{KPC}} & \multicolumn{2}{c}{\textbf{QRC}} \\
\cmidrule(r){1-1}\cmidrule(r){2-3}\cmidrule(r){4-5}
Metrics & Acc. & F1-macro & Accu. & F1-macro \\
\hline
TextCNN & 47.4 & 26.8 & 73.3 & 52.9  \\
TextRCNN & 55.3 & 38.8 & 79.6 & 59.0  \\
\hline
BERT & 59.6 & 34.9 & 82.7 & 63.4 \\
RoBERTa-wwm & 61.0 & 37.0 & 84.2 & 65.2  \\
Mengzi & 56.6 & 29.5 & 81.7 & 62.8  \\
BART & 62.7 & 41.9 & 82.0 & 63.0 \\
CPT & 66.2 & 48.4 & 82.8 & 63.4  \\
\hline
DAPT-BERT & 68.7 & 46.5 & 86.5 & 68.5  \\
MathBert & 68.9 & 47.1 & 85.3 & 69.8  \\
COMUS & 71.0 & \textbf{63.3} & 88.0 & 73.3 \\
DAPT-CPT & 72.0 & 58.0 & 88.8 & 76.7 \\
JiuZhang & \underline{73.3} & 59.4 & \underline{89.4} & \underline{79.2}  \\
JiuZhang-MTDNN & 71.5 & 58.4 & 89.2 & 77.1 \\
\hline
JiuZhang 2.0 (w/o IRL)& \textbf{73.5} & \underline{61.2} & \textbf{89.9} & \textbf{79.8}  \\
\bottomrule
\end{tabular}
\label{tab-main-results}
\end{table}

Note that CoT methods rely on  intermediate reasoning steps of the sampled exemplars in input to guide the solving of math problems, which are not available in the two classification tasks of KPC and QRC. While, in MCQ, BFQ, CAG and BAG tasks, we can utilize the analysis text to derive the intermediate reasoning steps, hence we only report the results of CoT methods on the four tasks.  

\paratitle{Implementation Details}. 
For GPT3-CoT and CodeX-CoT, we follow the standard chain-of-thought way to construct the input prompts~\cite{Wei2022ChainOT}, and the numbers of sampled exemplars are set to 5 and 8, respectively,  since GPT-3 has a smaller maximum input length than CodeX.
During training, we use AdamW~\cite{loshchilov2018decoupled} as the optimizer with the learning rate of $3\text{e-}5$, and warm up the learning rate for the first 5\% steps then decay the weight with a ratio of 0.01.
The coefficients of the auxiliary loss (Eq.~\eqref{eq-auxiliary-1}) and the $Z$-loss  (Eq.~\eqref{eq-auxiliary-3}) are $1\text{e-}3$ and $1\text{e-}4$, respectively. 
For the MoE structure, we set the number of experts $K=4$ and the number of activated experts $k=1$.
For continual multi-task pre-training, we pre-train our model with a batch size of 256 for 700000 steps.  
For multi-task fine-tuning, we fine-tune our model with a batch size of 32 for 80 epochs and adopt the routing mechanism with task prompt.
For iterative refinement, we use CodeX~\cite{chen2021evaluating} as the LLM and retrieve top-$8$ similar problems from the training set as exemplars for each input problem. 
More details are reported in Appendix~\ref{imp}.



\subsection{Main Results}
\label{sec:main-exp}

\subsubsection{Evaluation on Seen Tasks}\label{ex-seen}
For seen tasks, we evaluate the performance of our approach after multi-task fine-tuning.  The results of the seen QA/generation and classification tasks are shown in Table~\ref{tab-hard-results} and Table~\ref{tab-main-results}, respectively, and  we can observe that: 

First, continual pre-training methods (\ie COMUS, DAPT-CPT, JiuZhang, JiuZhang-MTDNN) achieve better performance than general-purpose PLMs such as BART and CPT. 
The reason is that these methods have been continually pre-trained on the math corpus, which can learn useful mathematical knowledge from such texts.
Among these continual pre-training methods, {the two methods based on JiuZhang (\ie JiuZhang and JiuZhang-MTDNN) mostly outperform all other methods. It is mainly because that  JiuZhang incorporates three types of pre-training tasks, which is further pre-trained in a curriculum learning way. While,  
JiuZhang-MTDNN revises the fine-tuning process of JiuZhang by adopting multi-task learning, which can improve the performance on MCQ and BFQ, but has worse performance on KPC and QRC tasks.
A possible reason is that there exists negative interference among these tasks during multi-task learning.}
Besides, COMUS also performs well on the KPC task.
Since the KPC task requires a deep  understanding of the formulas in  mathematical problems for predicting the knowledge points, COMUS specially designs graph neural networks and memory networks for modeling the formulas. 

Second, the chain-of-thought methods based on powerful LLMs (\ie GPT3-CoT and CodeX-CoT) overall perform worse than continual pre-training methods on generation metrics (\ie BLEU-4, ROUGE-2 and ROUGE-L). 
The reason might be that these LLMs mainly focus on English tasks, and cannot well adapt to Chinese math-related tasks.
In contrast, these continual pre-training methods have been trained over the math corpus, thus having an adaptation capacity in downstream tasks. 
While, for the {Accuracy} metric, chain-of-thought methods perform relatively better than other baselines. 
It shows that LLMs are more skilled in accurately predicting the answer, since they have a
stronger mathematical reasoning capacity due to the huge model size and large-scale pre-training corpus (also including large amounts of mathematical texts).

\begin{table*}
\centering
\caption{Main results on two analysis generation tasks for junior high school in the unseen  setting. }
\begin{tabular}{lcccccccc}
\bottomrule
\multirow{2}{*}{\textbf{Methods}} & \multicolumn{4}{c}{\textbf{JCAG}} & \multicolumn{4}{c}{\textbf{JBAG}} \\
\cmidrule(r){2-5}\cmidrule(r){6-9}
 & BLEU-4 & ROUGE-2 & ROUGE-L & Accuracy & BLEU-4 & ROUGE-2 & ROUGE-L & Accuracy \\
\hline
BART & 50.50 & 59.67 & 73.15 & 50.40 & 54.54 & 60.75 & 74.51 & 30.60 \\
CPT & 49.38 & 59.27 & 72.91 & 48.20 & 53.50 & 60.32 & 74.23 & 27.60 \\
DAPT-CPT & 52.06 & 60.84 & 73.53 & 54.50 & 54.66 & 61.36 & 74.78 & 32.30 \\
JiuZhang & 52.13 & 61.43 & 73.87 & 55.30 & 55.69 & 61.73 & 75.00 & 34.50 \\
\hline
JiuZhang 2.0 w/o IRL & \underline{53.37} & \underline{61.74} & \underline{74.00} & \underline{55.60} & \textbf{56.19} & \underline{62.13} & \underline{75.36} & \underline{38.10} \\
JiuZhang 2.0 & \textbf{55.73} & \textbf{63.76} & \textbf{75.37} & \textbf{63.20} & \underline{54.45} & \textbf{64.81} & \textbf{77.14} & \textbf{53.80} \\

\bottomrule
\end{tabular}
\label{tab-ood-results}
\end{table*}



Finally, our proposed JiuZhang 2.0 outperforms all the baselines in most cases. 
By integrating the MoE architecture with multi-task training, our model can better  capture the 
 mathematical knowledge across various math-related tasks. 
Even without iterative refinement via the LLM, our model  (\ie \emph{JiuZhang 2.0 w/o IRL}) can still outperform all the baselines.  
After incorporating the iterative refinement via the LLM, the performance of our approach can be further improved, especially on the Accuracy metric. 
It demonstrates that our approach can further benefit from the mathematical reasoning capacity of the LLM.
In this way, JiuZhang 2.0 can combine both the advantages of the PLM and LLM:  PLM can be tuned for domain adaptation to Chinese math-related tasks, while LLM has  stronger reasoning and generation capacities. 



\begin{figure*}[t!]
    \centering
    \includegraphics[width=\linewidth]{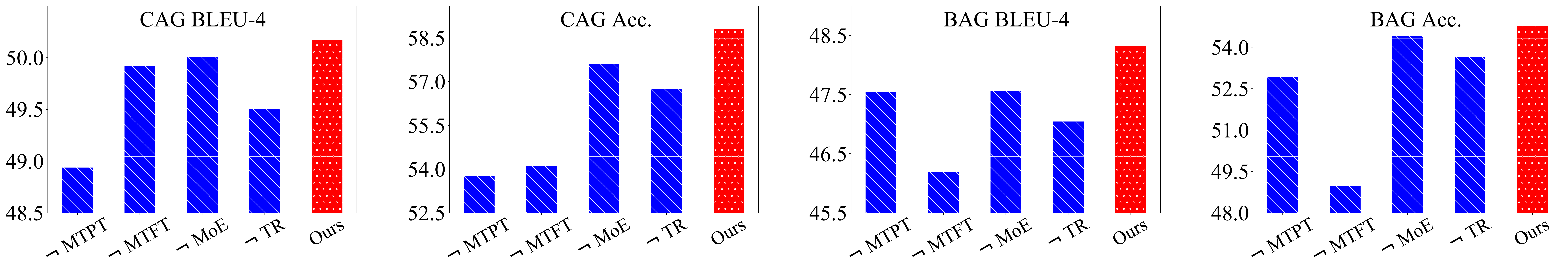}
    \caption{Ablation study of our approach on CAG and BAG tasks. 
    ``$\neg$'' indicates that the corresponding technique is removed from our model, while the rest are kept. We abbreviate the terms   Multi-task Continual Pre-Training, Multi-Task Fine-Tuning, Mixture-of-Experts, and Task embedding in Routing network as MTPT, MTFT, MoE and TR respectively.}
    \label{fig-ablation}
\end{figure*}

\subsubsection{Evaluation on Unseen Tasks}

Since multi-task fine-tuning cannot cover all math-related tasks, we continue to examine the performance of our model on new tasks that are not seen before.   
In order to enlarge the domain gap between existing and new tasks, we select the two tasks of multiple-choice analysis generation (JCAG) and blank-filling analysis generation (JBAG) from \emph{junior high schools}, which has a different distribution with those from \emph{high schools} (in multi-task fine-tuning).   
For these two unseen tasks, we fine-tune our model (task by task) on them after multi-task fine-tuning, as the same way in the baselines. 

From Table~\ref{tab-ood-results},  we can see that the overall experimental findings are similar to those discussed in Section~\ref{ex-seen}, where we have the overall performance order:  {PLMs} $<$ {continual pre-training methods} $<$ {JiuZhang} $<$ {JiuZhang 2.0 w/o IRL} $<$ {JiuZhang 2.0}.  
In particular, the variant of \emph{JiuZhang 2.0 w/o IRL} also performs better than all these baselines, since it employs MoE extension with multi-task training, thus having an improved ability for capturing common mathematical knowledge across tasks. 
Further,  by adopting the iterative refinement via LLMs~(IRL), our JiuZhang 2.0 achieves a 
significant improvement on the Accuracy metric (\ie $55.60\rightarrow63.20$ on JCAG, $38.10\rightarrow53.80$ on JBAG).  
The results show that the proposed IRL strategy can effectively leverage the strong  generation and reasoning  capacities of LLMs via in-context learning, which can gradually improve the generation quality of our PLM. 



\ignore{
Finally, by adopting the iterative refinement via LLMs~(IRL), our JiuZhang 2.0 consistently outperforms all other methods.
It further shows the effectiveness of using LLMs for improving the quality of our generated results.
Besides, we can see a significant improvement in the accuracy metric (\ie $55.60\longrightarrow63.20$ on JCAG, $38.10\longrightarrow53.80$ on JBAG).
The results show that the proposed IRL strategy can effectively benefit from the strong accurate generation capacity of LLMs, where the iterative usage of specially designed instructions can guide LLM to progressively correct possible errors in the generated results.
}

\subsection{Detailed Analysis}

\subsubsection{Ablation Study}
\label{sec:ablation}

In JiuZhang 2.0, we have proposed a series of  improvement techniques for enhancing  the capacity for mathematical problem solving. Next, we study how each technique contributes to the model performance. We keep the complete model with all improvement techniques as a reference,  then remove one specific technique each time, and compare the performance \emph{with} and \emph{without} it.    We consider the following variants: (1) 
\emph{$\neg$ MoE} removes the MoE extension, (2) \emph{$\neg$ MTPT} removes multi-task continual pre-training, (3) \emph{$\neg$ MTFT} removes multi-task fine-tuning, and (4) \emph{$\neg$ TR} removes the task embedding from the routing network. 
Note that \emph{$\neg$ MoE} can be considered as an implementation of the multi-task learning method~\cite{Liu2019MultiTaskDN} with JiuZhang as the backbone model. 
\ignore{Our proposed JiuZhang 2.0 incorporates MoE layers in the model architecture and adopts the multi-task pre-training and multi-task fine-tuning strategies to learn the model parameters.
And we also add task embeddings into the MoE layers to better guide the selection of experts according to the current task.
To verify the effectiveness of the above components, we conduct an ablation study that proposes four variations of our approach by removing the above components respectively, namely ours $\neg$ CPT, ours $\neg$ MTFT, ours $\neg$ MoE, ours $\neg$ TR.} 
We report BLEU-4 and Accuracy of these variants  on the CAG and BAG tasks. 

From Figure~\ref{fig-ablation}, we observe that removing any of these improvements  would lead to performance degradation, which  indicates the effectiveness of these proposed techniques in mathematical problem solving.   
In particular,  the removal of  multi-task pre-training or fine-tuning leads to a larger performance drop, which shows the two training strategies are more important to improve the model performance. These two tasks are well suited to the MoE architecture, and they can help capture the mathematical knowledge via the expert networks.

\begin{figure}
    \centering
    \includegraphics[width=0.45\textwidth]{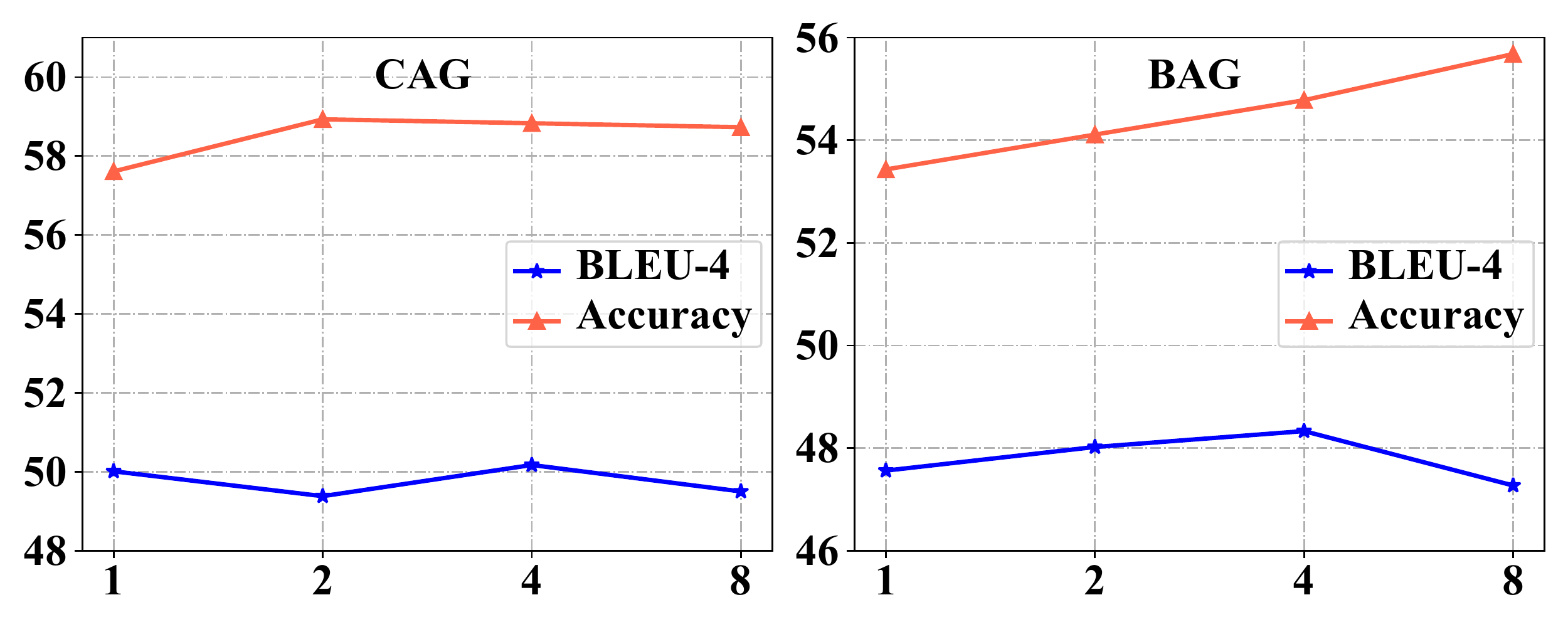}
    \caption{Varying the number of experts ($K$) in our approach. }
    \label{hyper_num}
\end{figure}

\begin{figure}
    \centering
    \includegraphics[width=0.45\textwidth]{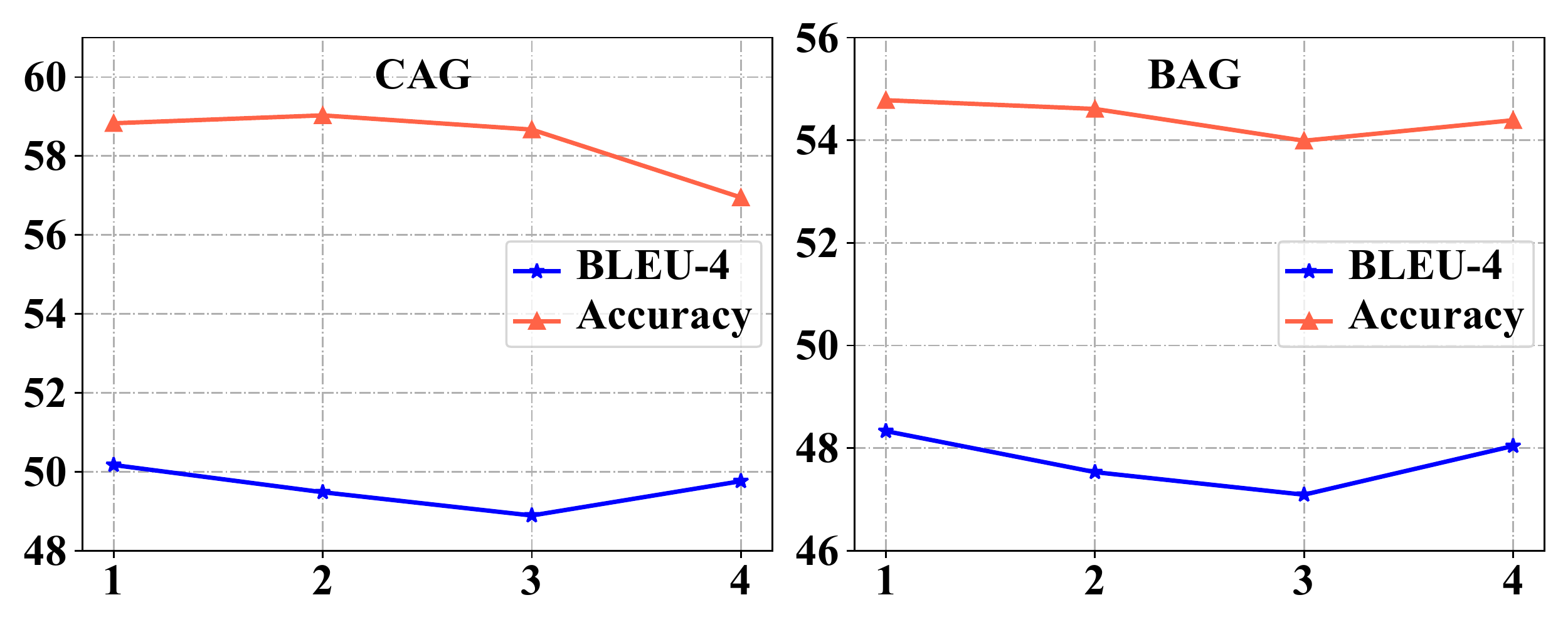}
    \caption{Varying the number of activated experts ($k$). }
    \label{hyper_top}
\end{figure}

\ignore{
\begin{figure}
    \centering
    \includegraphics[width=0.45\textwidth]{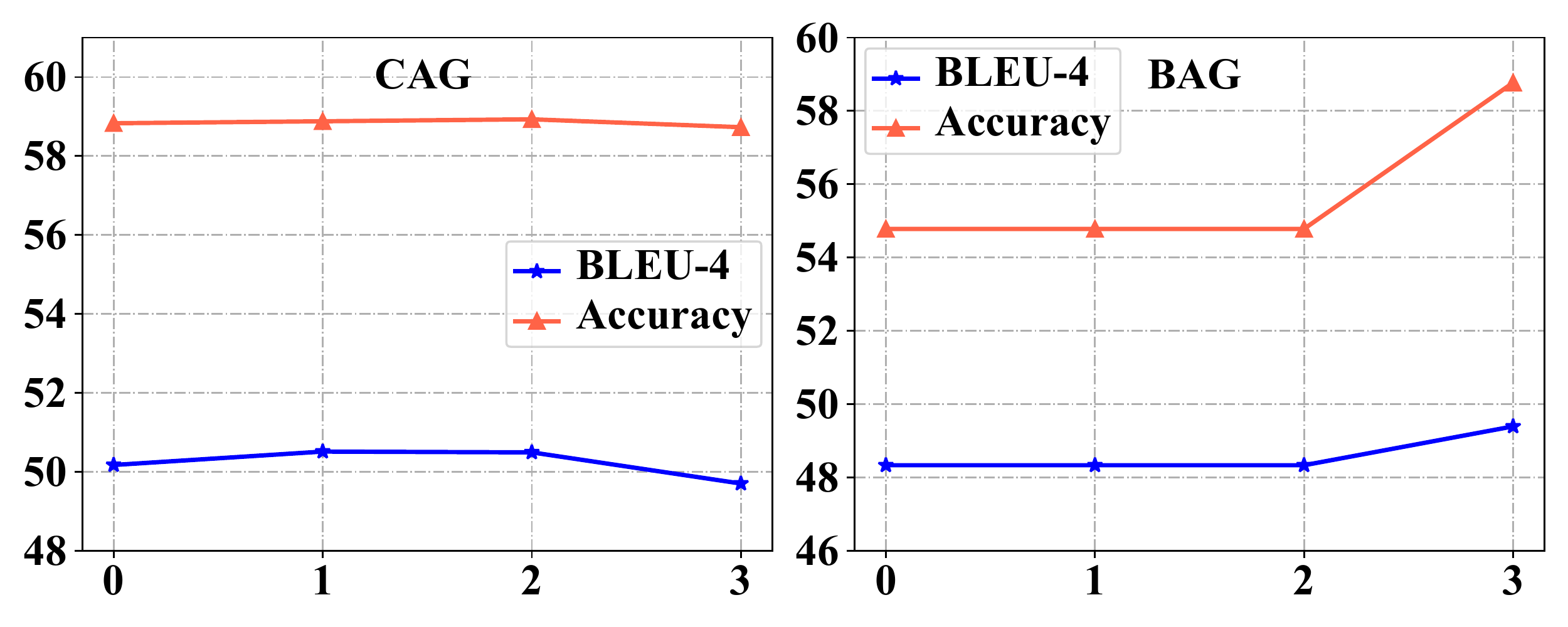}
    \caption{Performance comparison of our approach  \emph{w.r.t.} the number of iteration steps in each stages ($T$). }
    \label{hyper_ICL}
\end{figure}
}

\subsubsection{Hyper-parameters Analysis}
    

In our MoE architecture, there are two major hyper-parameters to tune,  \ie the number of experts $K$ and the number of activated experts $k$ in the MoE layers. 
Next, we investigate the effect of each hyper-parameter on our approach. 
We conduct the analysis experiments on CAG and BAG tasks and report the results on BLEU-4 and Accuracy metrics for the two hyper-parameters in Figure~\ref{hyper_num} and  Figure~\ref{hyper_top}, respectively.

 First, the increase  in the number of experts does not necessarily improve the performance of our approach (Figure~\ref{hyper_num}), especially in the Accuracy metric.
A possible reason is that the MoE architecture introduces additional parameters, which is more likely to overfit on the training set. Besides, using more experts also leads to larger computational costs. 
 In our experiments, to balance the effectiveness and efficiency, we set $K=4$, \ie using four expert networks, which generally gives a good performance.  
Second,  more activated experts are not useful  to improve the model performance, even leading to performance degradation (Figure~\ref{hyper_top}).
A possible reason is that activating more experts would cause interference among them, resulting in the conflict utilization of experts.
In contrast, by setting $k=1$, we can not only achieve a relatively better performance, but also save the computation cost of activated expert networks.

\ignore{Third, from Figure~\ref{hyper_ICL}, we can see that with the increasing of the iteratively refining steps, the performance of the model mostly increases.
It shows that more iterative turns can better leverage LLMs to improve the quality of the generated results.
Besides, we observe that the performance gain on the BAG task is more significant than on CAG.
The reason may be that the BAG task aims to generate proper analysis and answers into the blank, which is more difficult than CAG and requires more iterative turns to refine the generated analysis for obtaining the correct answer.}

\subsubsection{Analysis on the MoE Architecture}
A major contribution of our model lies in the architecture extension with MoE. 
By setting multiple expert networks, we can effectively share the mathematical knowledge  learned from the math corpus across tasks, so as to improve multi-task  mathematical problem solving.  These experts are expected to  capture  and decompose specific  mathematical knowledge for different math tasks. Next, we present an analysis experiment about the encoded knowledge at each expert network. 

As shown in Table~\ref{table-expert}, we select three mathematical texts from two tasks, and show the routed expert for each token (toke-level routing) in different background colors.   
It can be observed that our routing network can effectively decompose the mathematical  knowledge and route them to the  corresponding experts.
For example,  the trigonometric functions (\eg $sin$ and $\pi$) are  routed to \emph{expert \#3},  while the (background or formal) words and numbers are  mainly assigned to \emph{expert \#1} and \emph{expert \#2}, respectively. 

\ignore{It indicates that in the MoE layer, expert 3 has been selected to mainly focus on the knowledge of trigonometric functions, showing the effectiveness of our routing mechanism for decomposing different knowledge into corresponding experts.
Besides, we can see that the words and numbers are also mostly assigned to expert 1 and 2, respectively.
It shows that our approach can also decompose the capacity of understanding different tokens.}


\subsection{Online $A/B$ Test}

\begin{table}
\vspace{-0.2cm}
\centering
\caption{Online $A/B$ test of  JiuZhang 2.0 and JiuZhang via the automatic math problem solving function on Zhixuewang.}
\begin{tabular}{lcc}
\bottomrule
& JiuZhang 2.0 Wins & JiuZhang Wins \\
\hline
Ratio & 53.5 \% & 46.5\% \\
\bottomrule
\end{tabular}
\label{table-online}
\end{table}


Besides offline evaluation, we further conduct the online $A/B$ test on Zhixuewang\footnote{\url{https://www.zhixue.com/}} for examining the practical performance of our approach. Zhixuewang is designed as a teacher assistant app that provides personalized education services to students, accumulating about 51 million users in China mainland. 
Specially, we employ the function of \emph{automatic math problem solving} on Zhixuewang for conducting online $A/B$ test.  
Given a math problem (\eg  blank-infilling problem), this function aims to automatically generate the answer with a detailed analysis of the solving process.
Here, we compare our JiuZhang 2.0 with the original JiuZhang~\cite{zhao2022jiuzhang}, and both models are fine-tuned by the training data provided by this app.
For comparison, we sample a small population of requests of this function, and a user will be asked to select her/his preferred answer and analysis provided by the two models in each request.

Table~\ref{table-online}  reports the  winning ratio of the two methods.
As we can see, our proposed JiuZhang 2.0 performs better than the baseline JiuZhang. 
The major reason is that our model adopts the multi-task training with MoE layers to better capture the shared knowledge across multiple math-related tasks, and also leverages LLMs to iteratively refine  the generated results.
In this way, our model can generate more accurate answers and high-quality analysis.

\definecolor{e1}{RGB}{204, 255, 204}
\definecolor{e2}{RGB}{255, 214, 214}
\definecolor{e3}{RGB}{230, 230, 230}

\begin{CJK*}{UTF8}{gbsn}
\begin{table}
\small
\centering
\caption{Case study on the token-level expert routing. We use the background color to indicate different experts: \colorbox{e1}{{expert \#1}}, \colorbox{e2}{{expert \#2}}, \colorbox{e3}{{expert \#3}}.}
\label{cs-expert}
\begin{tabular}{cc}
\bottomrule
\textbf{Task} & \textbf{Problem Content} \\
\hline
BAG & \multicolumn{1}{c}{\begin{tabularx}{0.40\textwidth}{@{}X@{}}
  \sethlcolor{e2}\hl{$5$~}\sethlcolor{e3}\hl{$\sin 90^{\circ}~+$~}\sethlcolor{e2}\hl{$2$~}\sethlcolor{e3}\hl{$\sin$~}\sethlcolor{e2}\hl{$0^\circ -3$~}\sethlcolor{e3}\hl{$\sin$~}\sethlcolor{e2}\hl{$270^\circ$~}\sethlcolor{e3}\hl{$+$~}\sethlcolor{e2}\hl{$10$~}\sethlcolor{e3}\hl{$\cos$~}\sethlcolor{e2}\hl{$180$~}\sethlcolor{e3}\hl{$^\circ$~}\sethlcolor{e2}\hl{$=$}\sethlcolor{e3}\hl{\_\_\_\_ .}
  \end{tabularx}} \\
\hline
KPC & \multicolumn{1}{c}{\begin{tabularx}{0.40\textwidth}{@{}X@{}}
  \sethlcolor{e3}\hl{Known~}\sethlcolor{e2}\hl{the domain of definition of~}\sethlcolor{e3}\hl{function~}  \\ \sethlcolor{e2}\hl{$y$~}\sethlcolor{e3}\hl{$=$~}\sethlcolor{e2}\hl{$2$}\sethlcolor{e3}\hl{$a$}\sethlcolor{e3}\hl{$\cos$}\sethlcolor{e2}\hl{$(2$}\sethlcolor{e3}\hl{$x$}\sethlcolor{e2}\hl{$-$}$\frac{\colorbox{e3}{$\pi$}}{\colorbox{e2}{$3$}}$\sethlcolor{e3}\hl{$)$}\sethlcolor{e2}\hl{$~+~b$~}\sethlcolor{e2}\hl{is~}\sethlcolor{e3}\hl{$[$}\sethlcolor{e2}\hl{$0$}\sethlcolor{e3}\hl{$,$~}$\frac{\colorbox{e3}{$\pi$}}{\colorbox{e2}{$2$}}$\sethlcolor{e3}\hl{~$]$~}\sethlcolor{e2}\hl{and the domain of function is~}\sethlcolor{e2}\hl{[$-5$}\sethlcolor{e3}\hl{$,$~}\sethlcolor{e2}\hl{1}\sethlcolor{e3}\hl{$]$.~}\sethlcolor{e3}\hl{Find the value of~}\sethlcolor{e3}\hl{$a$,~}\sethlcolor{e2}\hl{$b$}\sethlcolor{e3}\hl{.}
  \end{tabularx}} \\
\hline
KPC & \multicolumn{1}{c}{\begin{tabularx}{0.40\textwidth}{@{}X@{}}
  \sethlcolor{e3}\hl{A seagoing ship~}\sethlcolor{e1}\hl{starts from~}\sethlcolor{e3}\hl{A}\sethlcolor{e1}\hl{,~}\sethlcolor{e3}\hl{sails~}\sethlcolor{e1}\hl{in a straight line~}\sethlcolor{e1}\hl{at a speed of~}\sethlcolor{e1}\hl{$40$ nautical miles per hour~}\sethlcolor{e1}\hl{in the direction of~}\sethlcolor{e2}\hl{$40$}\sethlcolor{e1}\hl{$^\circ$~} ......
  \end{tabularx}} \\
\bottomrule
\end{tabular}
\label{table-expert}
\end{table}
\end{CJK*}

\section{Conclusion}
In this paper, we proposed JiuZhang~2.0, a unified  Chinese PLM for multi-task mathematical problem solving. Different from previous PLM approaches for math domain, we focus on improving the multi-task capacity of PLMs, especially on complex tasks.  
For this purpose,  we designed a MoE-based encoder for modeling the mathematical text, aiming to  share the mathematical knowledge across different tasks.  
To support the MoE architecture, we specially designed multi-task continual pre-training and multi-task fine-tuning strategies for learning the shared knowledge via  expert networks. Further, we leveraged the powerful LLMs as a complementary role to iteratively refine the generation results by our PLM, with the elaborately designed prompts. 
Experimental results (both offline evaluation and online $A/B$ test) have demonstrated that our approach is superior to competitive baselines on a variety of math-related tasks.  


\section*{Acknowledgement}
This work was partially supported by National Natural Science Foundation of China under Grant No. 62222215, Beijing Natural Science Foundation under Grant No. 4222027, and Beijing Outstanding Young Scientist Program under Grant No. BJJWZYJH012019100020098. And this work is also partially supported by the Outstanding Innovative Talents Cultivation Funded Programs 2021 of Renmin University of China. Xin Zhao is the corresponding author.
\ignore{
This work was partially supported by Beijing Natural Science Foundation under Grant No. 4222027, and  National Natural Science Foundation of China under Grant No. 61872369, Beijing Outstanding Young Scientist Program under Grant No. BJJWZYJH012019100020098,
and the Outstanding Innovative Talents Cultivation Funded Programs 2021. This work is also partially supported by Beijing Academy of Artificial Intelligence~(BAAI). Xin Zhao is the corresponding author.
}
\bibliographystyle{ACM-Reference-Format}
\balance
\bibliography{sample-base}

\clearpage
\newpage
\renewcommand{\appendixpagename}{Supplementary material}
\begin{appendices}

\ignore{
\section{Pre-training Tasks}
\label{pt_task}
In the multi-task pre-training stage, we reuse the original three types of pre-training tasks of JiuZhang~\cite{zhao2022jiuzhang}, and re-organize them into a multi-task pre-training objective.
Here, we introduce the three types of tasks in detail.

\paratitle{Basic Course: Masked Token Prediction.}
The basic course aims to enhance the basic understanding of math symbols and establish semantic connections between math symbols and textual words. Given a mathematical text, 15\% tokens of the input sequence are selected for masking (including both words and symbols), in which 80\% ones are replaced by the token ``\textsf{[MASK]}'', 10\% ones are replaced by a random token and the rest 10\% ones remain unchanged. Let $\widetilde{x}$ denote the masked sequence,  $V_{mask}$ denotes the selected tokens, and the MLM and DAE losses are defined as:
\begin{align}
     L_{MLM} &=\sum_{t_i \in V_{mask}}-\log{p(t_i | \widetilde{x};\Theta_{E}, \Theta_{U})},
    \label{eq-mlm}\\
     L_{DAE} &=\sum_{i}-\log{p(t_i|t_{<i}, \widetilde{x};\Theta_{E}, \Theta_{G})},
    \label{eq-dae}
\end{align}
where $p(t_i | \widetilde{x};\Theta_{E}, \Theta_{U})$ and $p(t_i|t_{<i};\widetilde{x};\Theta_{E}, \Theta_{G})$ denote the prediction probabilities of the token $t_i$ at the $i$-th position according to $U$-decoder and $G$-decoder, respectively.
Besides, based on the causality of math solutions, we also devise a linearly increasing weighting mechanism that assigns larger masked probability at larger positions.

\ignore{
Based on the causality of math solution, a linearly increasing weighting mechanism that assigns larger masked probability in larger positions is utilized in masking tokens:
\begin{equation}
    s_i = \frac{30}{n-1} \times i,
    \label{eq-mask}
\end{equation}
where $ s_i$ is the sampling weight to be masked for the $i$-th position, $n$ is the sentence length, and $i$ is the position index ranging from 0 to $n-1$.} 

\paratitle{Advanced Course: Mathematical Logic Recovering.}
To fully comprehend mathematical problems and the logical derivations necessary, logic-based pre-training tasks are proposed with the reconstruction of shuffled sentences and formulas. 
For the Shuffled Sentences Recovering (SSR) task, the sentences in the solution text (denoted as $d$) are shuffled, producing a corrupted solution text denoted as $\widetilde{d}_{S}$. The pre-training objective is to recover the original solution text $d$ based on $\widetilde{d}_{S}$ with $G$-decoder:
\begin{equation}
    L_{SSR}=\sum_{i}-\log{p(t_i|t_{<i},  \widetilde{d}_{S};\Theta_{E}, \Theta_{G}} ),
    \label{eq-ssr}
\end{equation}
For the Shuffled Formulas Recovering~(SFR) task, the formulas $\{f_1, f_2, \cdots , f_m\}$ from the solution text are shuffled to construct the corrupted solution text denoted as $\widetilde{d}_{F}$. The PLM aims to recover the original text based on $\widetilde{d}_{F}$:
\begin{equation}
    L_{SFR}=\sum_{i}-\log{p(t_i|t_{<i}, \widetilde{d}_{F};\Theta_{E}, \Theta_{G}}).
    \label{eq-sfr}
\end{equation}

\paratitle{Advanced Course: Solution Checking.}
To improve the sensitivity of PLMs to solution errors, the Dual-Decoder Solution Checking~(SC) task is adopted as an advanced course. First, the two decoders are utilized to fill the masked symbols or words. Then, they are trained to detect and correct the solution generated by each other.
Specifically, given the solution text $d=\{t_1,t_2,\cdots,t_l\}$, a proportion of tokens following the same strategy in the basic course are selected to produce the masked text $\widetilde{x}$.
Then, the $U$-decoder and $G$-decoder generate recovered results, $\widetilde{d}_{U}$ and $\widetilde{d}_G$, based on their own generation probabilities of the masked tokens.
Next, the generated texts $\widetilde{d}_{G}$ and $\widetilde{d}_U$ are then evaluated and corrected by the $U$-decoder and $G$-decoder. Therefore, errors in generations will be detected and corrected through a dual process. Such a pre-training process can be formulated as:
\begin{align}
    L_{USC} &= \sum_{i}-\log{p(t_i|\widetilde{d}_G;\Theta_{E},\Theta_{U})}, \label{eq-usc}\\
    L_{GSC} &=\sum_{i}-\log{p(t_i|t_{<i},
    \widetilde{d}_U;\Theta_{E},\Theta_{G})}, 
    \label{eq-gsc}
\end{align}
where $L_{USC}$ and $L_{GSC}$ denote the self-correction losses for the $U$-decoder and $G$-decoder, respectively.}

\begin{table}[t!]
\small
\centering
\caption{Parameter settings of our models.}
\label{tab-baseline}
\begin{tabular}{|c|c|} 
\hline
\textbf{Task} & \textbf{Settings}                    \\ 
\hline
\hline
Continual Pre-training & \begin{tabular}[c]{@{}c@{}}AdamW, learning\_rate=3e-5\\warmup\_ratio=0.01\\batch\_size=256\\max\_steps=70k\\num\_experts=4\\top\_k=1\end{tabular} \\
\hline
Multi-task Fine-tuning & \begin{tabular}[c]{@{}c@{}}AdamW, learning\_rate=3e-5\\warmup\_ratio=0.1\\batch\_size=64\\num\_experts=4\\top\_k=1\\router=task\_router\end{tabular} \\
\hline
OOD Fine-tuning & \begin{tabular}[c]{@{}c@{}}AdamW, learning\_rate=5e-5\\warmup\_ratio=0.1\\batch\_size=64\\num\_experts=4\\top\_k=1\\router=task\_router\end{tabular} \\ 
\hline
In-context Learning & \begin{tabular}[c]{@{}c@{}}num\_examplars=8\\
$T$=1\\\end{tabular} \\
\hline
\end{tabular}
\end{table}

\begin{algorithm}
  \caption{The multi-task training algorithm.}\label{al_moe}
\small
    \SetKwData{Left}{left}\SetKwData{This}{this}\SetKwData{Up}{up}
    \SetKwFunction{Union}{Union}\SetKwFunction{Sample}{Sample}\SetKwFunction{MaskMLM}{MaskMLM}\SetKwFunction{MaskDAE}{MaskDAE}
    \SetKwInOut{Input}{Input}\SetKwInOut{Parameter}{Parameter}

    \Input{Pre-training corpus, Multiple math-related datasets for fine-tuning}
    \Parameter{The parameters of the encoder $\Theta_{E}$, $U$-decoder $\Theta_{U}$, $G$-decoder $\Theta_{G}$}
    \BlankLine
    \tcp{Multi-task Pre-training for MoE Adaptation}
    \While{not converged}{
        Sample a batch from the pre-training corpus\;
        Compute the six pre-training losses using Eq.~\ref{eq-continue-pretrain}\;
        Compute the two auxiliary losses using Eq.~\ref{eq-auxiliary-1} and Eq.~\ref{eq-auxiliary-3}\;
        Performing gradient descent to optimize $\Theta_{E}$, $\Theta_{U}$ and $\Theta_{G}$\;
    }

    \tcp{Multi-task Fine-tuning for MoE Adaptation}
    \While{not converged}{
        Sample a batch from the multiple fine-tuning datasets\;
        Unify the input and output data formats of the batch of instances\;
        Compute the fine-tuning loss using the $U$-decoder and $G$-decoder\;
        Performing gradient descent to optimize $\Theta_{E}$, $\Theta_{U}$ and $\Theta_{G}$\;
    }
    
\end{algorithm}\DecMargin{1em}

\begin{algorithm}
  \caption{The iteratively refining algorithm.}\label{al_icl}
\small
    \SetKwData{Left}{left}\SetKwData{This}{this}\SetKwData{Up}{up}
    \SetKwFunction{Union}{Union}\SetKwFunction{Sample}{Sample}\SetKwFunction{MaskMLM}{MaskMLM}\SetKwFunction{MaskDAE}{MaskDAE}
    \SetKwInOut{Input}{Input}\SetKwInOut{Parameter}{Parameter}

    \Input{A mathematical problem $q$ and its analysis generated by JiuZhang with MoE $\tilde{a}^{(0)}$}
    \Parameter{Iteration steps of each stage $T$, number of exemplars $B$}
    \BlankLine
    \For{$s\leftarrow 1$ \KwTo $3$}{ 
        \For{$t\leftarrow 1$ \KwTo $T$}{
            $\text{iter\_step} \leftarrow (s - 1) \times T + t$ \;
            $\hat{a}^{(\text{iter\_step})} \leftarrow \tilde{a}^{(\text{iter\_step} - 1)}$ \;
            \Switch{$s$}{
                \Case{1}{
                    Use $q$ as the query for retrieval \;
                }
                \Case{2}{
                    Use $q$ and $\hat{a}^{(\text{iter\_step})}$ as the query for retrieval \;
                }
                \Case{3}{
                    Use $\hat{a}^{(\text{iter\_step})}$ as the query for retrieval \;
                }
            }
            Retrieve exemplars $C = \{ \langle q_j, a_j \rangle \}^B_{j = 1}$ from the training set using the query\;
            Construct input prompt using Eq.~\ref{eq-prompt}\;
            Feed the prompt to CodeX to obtain the refined result $\tilde{a}^{(\text{iter\_step})}$\;
        }
    }
\end{algorithm}\DecMargin{1em}

\section{Implementation Details.}
\label{imp}

We report the detailed parameter settings of our approach throughout the experiments in Table~\ref{tab-baseline}.
In addition to the above settings, we conduct all the experiments on 8 RTX 3090 24G GPUs, where the multi-task continual pre-training and multi-task fine-tuning took about 72 and 12 hours, respectively.
During multi-task fine-tuning, we construct the model inputs of all downstream tasks as follows, and the task embedding will be inserted after the $[\textsc{CLS}]$ token embedding.

\textbf{KPC, MCQ, BFQ, CAG, BAG, JCAG and JBAG:} [CLS] $q$ [SEP].

\textbf{QRC:} [CLS] $q_1$ [SEP] $q_2$ [SEP].

For the iterative refinement via the LLM, we also design three types of instructions and adopt three ways to construct queries for retrieval, for the three iterative stages, respectively.
We show the details in Table~\ref{iter-setting}.

Besides, we also present the Algorithm~\ref{al_moe} and Algorithm~\ref{al_icl}, to better show the multi-task training and iterative refinement processes of our approach, respectively. 

\section{Inference Latency Analysis}
\label{sec:latency}

\begin{table}
\centering
\caption{The inference latency per batch of different methods on BAG.}
\begin{tabular}{lccccc}
\bottomrule
\multirow{2}{*}{\textbf{Methods}} & \multicolumn{5}{c}{\textbf{BAG}} \\
& Latency  & BLEU-4 & Accuracy \\
\hline
BART & 830 ms & 45.46 & 43.92  \\
CPT & 330 ms & 44.82 & 40.68 \\
JiuZhang 2.0 w/o IRL & 370 ms & \textbf{48.33} & \textbf{54.78} \\
\bottomrule
\end{tabular}
\label{table-latency}
\end{table}

\begin{CJK*}{UTF8}{gbsn}
\begin{table*}
\renewcommand\arraystretch{1.2}
\centering
\caption{The detailed retrieved queries and instructions for constructing the input in each iterative refinement stage.}
\begin{tabular}{|l|l|l|}
 \hline
 \textbf{Iteration Stage} & \textbf{Query for Retrieval} & \textbf{Instruction} \\
 \hline
 \hline
 First Stage & Problem statement $q$ & \multicolumn{1}{l|}{\begin{tabularx}{0.5\textwidth}{@{}X@{}}
  根据上文中相似的题目，选取正确的解题思路，修改本题参考解答中的错误。 \\ According to similar problems above, choose the correct idea to solve this problem and fix mistakes in the reference solution.
  \end{tabularx}} \\
 \hline
 Second Stage & Problem statement $q$ + Generated analysis $\hat{a}$ & \multicolumn{1}{l|}{\begin{tabularx}{0.5\textwidth}{@{}X@{}}
  根据上文中相似的题目，修改本题的参考解答中的推理错误和逻辑错误，确定本题的答题逻辑，得到正确解答。 \\ According to similar problems above, correct the reasoning errors and logical errors in the reference solution, determine the reasoning logic of this question and get the correct solution.
  \end{tabularx}} \\
 \hline
 Third Stage & Generated analysis $\hat{a}$ & \multicolumn{1}{l|}{\begin{tabularx}{0.5\textwidth}{@{}X@{}}
  根据上文中相似的题目，修改本题的参考解答中的计算错误和抄写错误，确定本题的正确答案，得到正确解答。 \\ According to similar problems above, correct the calculation errors and transcription errors in the reference solution, determine the final answer and get the correct solution. 
  \end{tabularx}} \\
 \hline
\end{tabular}
\label{iter-setting}
\end{table*}
\end{CJK*}

\begin{CJK*}{UTF8}{gbsn}
\begin{table*}
\renewcommand\arraystretch{1.4} 
\caption{Case study on the analysis generation tasks.}
\begin{tabular}{|l|ll|}
\hline
\textbf{Math Problem} &
  \multicolumn{2}{l|}{\begin{tabularx}{0.85\textwidth}{@{}X@{}}
  设 $\{a_n\}$ 为等差数列 , $a_1 = 22$ , $S_n$ 为其前 $n$ 项和 , 若 $S_{10} = S_{13}$ , 则公差 $d = ?$  \\ 
  Let $\{a_n\}$ be an arithmetic sequence, $a_1 = 22$ , $S_n$ be the sum of the first $n$ items, if $S_{10} = S_{13}$ , the common difference $d = ?$ \\
  A. $- 2$ ; B. $- 1$ ; C. $1$ ; D. $2$ ;\\
  \end{tabularx}} \\ \hline
\textbf{Ground-Truth} &
  \multicolumn{2}{l|}{\begin{tabularx}{0.85\textwidth}{@{}X@{}}
  因为 $S_{10} = S_{13}$ , $a_1 = 22$ , 所以 $10 \times 22 +  \frac{10(10 - 1)}{ 2 } d = 13 \times 22 +  \frac{13(13 - 1)}{ 2 } d$ , 解得 $d = - 2$ . 故选 A . \\
  Because $S_{10} = S_{13}$ , $a_1 = 22$ , so $10 \times 22 + \frac{10(10 - 1)}{ 2 } d = 13 \times 22 + \frac{13( 13 - 1)}{ 2 } d$ , the solution is $d = - 2$ . So choose A .
  \end{tabularx}} \\ \hline
\textbf{Codex-CoT} &
  \multicolumn{2}{l|}{\begin{tabularx}{0.85\textwidth}{@{}X@{}}
由 $S_{10} = S_{13}$ 得到 $a_1 + \cdots + a_{10} = a_1 + \cdots + a_{13}$ , 即 $a_{13} = 0$. 因此 $a_{13} = a_1 + (13 - 1) d = a_1 - 12 d = 0$ , 解得 $d = 1$ . 故选 C . \\
  From $S_{10} = S_{13}$ , get $a_1 + \cdots + a_{10} = a_1 + \cdots + a_{13}$ , namely $a_{13} = 0$. Therefore $a_{13} = a_1 + (13 - 1) d = a_1 - 12 d = 0$ , the solution is $d = 1$ . So choose C.
  \end{tabularx}} \\ \hline
\textbf{JiuZhang} &
  \multicolumn{2}{l|}{\begin{tabularx}{0.85\textwidth}{@{}X@{}}
  由 $S_{10} = S_{13}$, 得 $\frac{10(a_1 + a_{13})}{2} = \frac{13(a_1 + a_{13})}{2}$, 即 $22 + 11 d = 0$, 解得 $d = 2$. 故选 D. \\
  From $S_{10} = S_{13}$, get $\frac{10(a_1 + a_{13})}{2} = \frac{13(a_1 + a_{13})}{2}$, That is, $22 + 11 d = 0$, the solution is $d = 2$. So choose D.
  \end{tabularx}} \\ \hline
\textbf{Ours} &
  \multicolumn{2}{l|}{\begin{tabularx}{0.85\textwidth}{@{}X@{}}
  因为 $S_{10} = S_{13}$ , 所以 $a_{11} + a_{12} + a_{13} = 0$ , 所以 $a_{12} = 0$ , 又 $a_1 = 22$ , 所以 $d = - 2$ . 故选 A . \\
  Because $S_{10} = S_{13}$ , so $a_{11} + a_{12} + a_{13} = 0$ , so $a_{12} = 0$ , and $a_1 = 22$ , So $d = - 2$ . So choose A .
  \end{tabularx}} \\ \hline 
\end{tabular}
\label{case-study}
\end{table*}
\end{CJK*}

In our approach, although we scale up the number of parameters in the PLM by incorporating the MoE layers, the sparsely routing mechanism can ensure that only the top-$1$ most related expert will be activated, leading to relatively less increased computational cost.
To investigate it, we conduct the analysis experiments to compare the inference latency per batch of our model with two baselines using different model structures, \ie BART and CPT, in the BAG task. 
During inference, we adopt greedy search to decode and set the batch size to 16. 
As shown in Table~\ref{table-latency}, compared to CPT, the inference latency of our model is slightly increased.
It indicates the effectiveness of the sparse routing mechanism to guarantee the efficiency of our approach.
Besides, we can see that BART requires double the inference time of CPT and our approach.
The reason is that CPT and JiuZhang 2.0 adopt an unbalanced model architecture with a shallower decoder than BART (2 layers VS. 6 layers), which can save the computation cost on the cross-attention layers of the decoder.

\section{Case Study}
To give a qualitative analysis of our proposed approach, we perform a case study that shows the generated analysis from our approach.
We select two examples from the CAG and BAG tasks, respectively, and also show the generated analysis by two best performed methods, \ie JiuZhang and CodeX-CoT.

As shown in Table~\ref{case-study}, although CodeX-CoT and JiuZhang have generated a detailed multi-step reasoning process consisting about the two problems, they both make mistakes in the intermediate steps.
For the first example, we can see that CodeX-CoT obtains a wrong intermediate conclusion $a_{13}=0$ by mistakenly simplifying the summation of two arithmetic progressions, which may be caused by the unfamiliarity of the knowledge about arithmetic progressions.
JiuZhang makes a small mistake in calculation, \ie $22+11d=0\longrightarrow d=2$, leading to the wrong answer.
It also reflects the lack of mathematical computation common sense about JiuZhang.
As a comparison, we can see that our approach can generate more proper analysis and successfully produce the true answers.
It indicates the effectiveness of our approach in solving complex mathematical problems.
\end{appendices} 

\end{document}